\documentclass[10pt,twocolumn,letterpaper]{article}

\usepackage[pagenumbers]{cvpr} % To force page numbers, e.g. for an arXiv version
\usepackage[accsupp]{axessibility}  % Improves PDF readability for those with disabilities.
\definecolor{cvprblue}{rgb}{0.21,0.49,0.74}
\usepackage[pagebackref,breaklinks,colorlinks,allcolors=cvprblue]{hyperref}
\usepackage{multirow}
\usepackage{graphicx}
\usepackage{xcolor}
\usepackage{utfsym}

\def\paperID{2403} % *** Enter the Paper ID here
\def\confName{CVPR}
\def\confYear{2025}

\title{Hybrid-Level Instruction Injection for Video Token Compression\\ in Multi-modal Large Language Models}

\author{
Zhihang Liu\textsuperscript{1}\thanks{Interns at Alibaba Group},
Chen-Wei Xie\textsuperscript{2},
Pandeng Li\textsuperscript{1,2}\thanks{Corresponding author},
Liming Zhao\textsuperscript{2},
Longxiang Tang\textsuperscript{3}, \\
Yun Zheng\textsuperscript{2},
Chuanbin Liu\textsuperscript{1},
Hongtao Xie\textsuperscript{1}
\\[0.4ex]
\textsuperscript{1~}University of Science and Technology of China\quad \\
\textsuperscript{2~}Tongyi Lab, Alibaba Group\quad
\textsuperscript{3~}Tsinghua University\quad \\
{\tt\small \{liuzhihang, lpd\}@mail.ustc.edu.cn, htxie@ustc.edu.cn}
}
\begin{document}
\maketitle
\begin{abstract}
Recent Multi-modal Large Language Models (MLLMs) have been challenged by the computational overhead resulting from massive video frames, often alleviated through compression strategies. However, the visual content is not equally contributed to user instructions, existing strategies (\eg, average pool) inevitably lead to the loss of potentially useful information. To tackle this, we propose the \textbf{H}ybrid-level Instruction \textbf{I}njection Strategy for \textbf{C}onditional Token C\textbf{om}pression in MLLMs (HICom), utilizing the instruction as a condition to guide the compression from both local and global levels. This encourages the compression to retain the maximum amount of user-focused information while reducing visual tokens to minimize computational burden. Specifically, the instruction condition is injected into the grouped visual tokens at the local level and the learnable tokens at the global level, and we conduct the attention mechanism to complete the conditional compression. From the hybrid-level compression, the instruction-relevant visual parts are highlighted while the temporal-spatial structure is also preserved for easier understanding of LLMs. To further unleash the potential of HICom, we introduce a new conditional pre-training stage with our proposed dataset HICom-248K. Experiments show that our HICom can obtain distinguished video understanding ability with fewer tokens, increasing the performance by 2.43\% average on three multiple-choice QA benchmarks and saving 78.8\% tokens compared with the SOTA method. The code is available at \url{https://github.com/lntzm/HICom}.
\end{abstract}

\section{Introduction}
\label{sec:intro}

\begin{figure}[!t]
\centering
\includegraphics[width=0.45\textwidth]{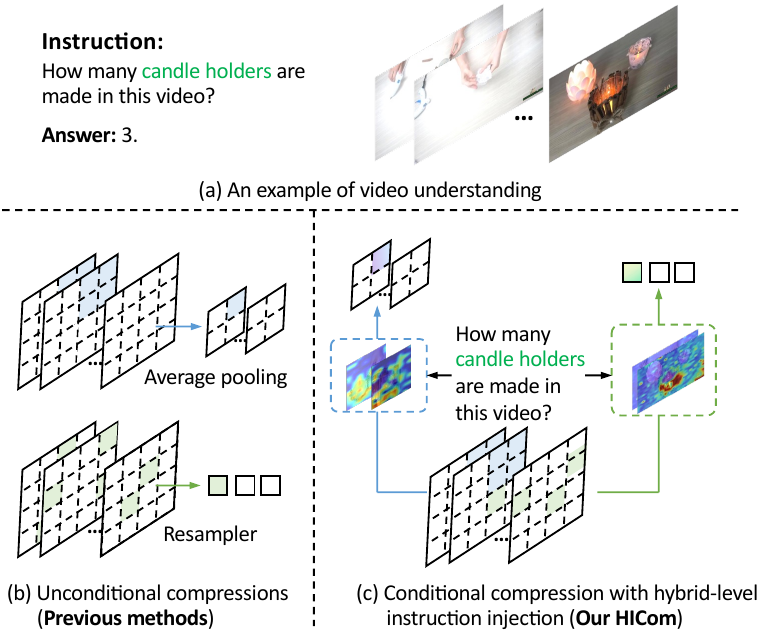}
\caption{An example of the video understanding task, and the comparison between the unconditional compression and our proposed conditional compression with hybrid-level instruction injection. We inject instruction at both local and global levels, guiding the compression to retain the maximum amount of user-focused information and minimize the computational burden.}
\label{fig:intro}
% \vspace{-8pt}
\end{figure}

Multi-modal Large Language Models~\cite{blip2, llava, llava1_5, minigpt4, qwenvl, ge2024alignzeg} (MLLMs) have gained significant improvements in multi-modal understanding by integrating visual information into the LLMs, beating various expert methods on downstream tasks~\cite{gao2024self, ge2024towards, xu2024fakeshield, zhang2024editguard, tang2024mind, li2023momentdiff, li2023progressive, zhang2024choose}.
While initially focused on image understanding, more research~\cite{videollava, videochatgpt, btadapter, wei2024dreamvideo, wei2024dreamvideo2, qu2025doesvisionlanguagemodellost} has shifted towards challenging video tasks.
Most MLLMs handle videos by treating them as sequences of images, sampling frames, and concatenating visual tokens from these frames~\cite{gemini1_5, internvl1_5, llavaov, qwen2vl, videollama3}. 

Compared to images, videos comprise multiple frames, resulting in more visual tokens and thus higher computational costs.
To make the computation affordable, early methods~\cite{videollava, videochat, btadapter} often employ sparse temporal sampling strategies, leading to significant loss of temporal information. Consequently, current MLLMs focus on compressing visual tokens to achieve a trade-off between computational costs and video understanding ability. Mainstream MLLMs~\cite{llavanextvideo, llavaov, pllava} use a spatial pooling strategy within each frame, leveraging capabilities gained from image data. 
Some approaches~\cite{videollama, otter, timechat} use Q-Former~\cite{blip2} or Resampler~\cite{flamingo} to compress the spatial-temporal information into a fixed number of visual tokens.
Other methods employ convolution~\cite{videollama2}, clustering~\cite{chatunivi}, or memory mechanisms~\cite{flashvstream, malmm, moviechat} to compress visual tokens in both spatial and temporal dimensions.
However, these compression methods are unconditional, lacking explicit guidance, as illustrated in \cref{fig:intro} (b), making it challenging to achieve high compression ratios with minimal information loss. Consequently, user-relevant visual content may be overlooked during such unconditional compression.

Therefore, we focus on introducing the instruction information as a condition to perform targeted compression, as it is rarely explored in previous work.
As shown in \cref{fig:intro} (c), conditional compression uses explicit instructional guidance to select user-focused information for retention, thereby enhancing efficiency. 
To thoroughly explore the conditional compression, we propose \textbf{H}ybrid-level Instruction \textbf{I}njection Strategy for \textbf{C}onditional Token C\textbf{om}pression in MLLMs (HICom).
Observing how humans process complex visual information, they usually adopt a coarse-to-fine approach, first considering the relative parts in the global coarse impression and then seeking detailed information locally in the video.
Motivated by this human perception process, HICom conducts the compression at both local and global levels, retaining the maximum amount of instruction-relevant information while preserving the temporal-spatial structure with fewer tokens, thus achieving a better balance between the computational burden and video comprehension.

Specifically, at the local level, we divide the sampled frame features into several groups and compress the tokens of each group into one token based on the attention with the injected instruction condition. Different from Q-Former~\cite{blip2} and Resampler~\cite{flamingo}, the grouped local attention can explicitly maintain the spatial-temporal structural characteristics of the video while highlighting the relevant parts within each group. 
At the global level, we inject the instruction condition into a small number of learnable tokens and then conduct the attention with the flattened frame features without grouping. The global compression focuses more on searching instruction-relevant information in the whole video and thus can be expressed by a small number of tokens as an auxiliary. 
To further unleash the potential of HICom, we introduce a new conditional pre-training stage between the alignment and instruction tuning and construct a new dataset HICom-248K of 248K video clips with high-quality instruction-followed descriptions. The proposed new stage continues to pre-train the parameters of condition injection, further improving the performance. Experiments show that our HICom can achieve state-of-the-art performance on 5 benchmarks with significantly fewer tokens. (\eg, increasing by 2.43\% average on three multiple-choice QA benchmarks and saving 78.8\% tokens compared with LLaVA-Video-7B~\cite{llavavideo}).

Our main contributions are listed as follows: 
\begin{itemize}
    \item We propose a conditional compression method HICom for MLLMs,  achieving effective video understanding while significantly reducing the computational burden.
    \item We conduct hybrid-level instruction injection to achieve the conditional compression, and introduce a new conditional pre-training stage with our constructed HICom-248K dataset to further unleash the potential of the conditional compressing.
    \item Experiments on various benchmarks show the proposed HICom can achieve better performance with fewer visual tokens, providing valuable insights for MLLMs.
\end{itemize}

\section{Related Work}
\label{sec:related_work}

\begin{figure*}[!t]
\centering
\includegraphics[width=0.85\textwidth]{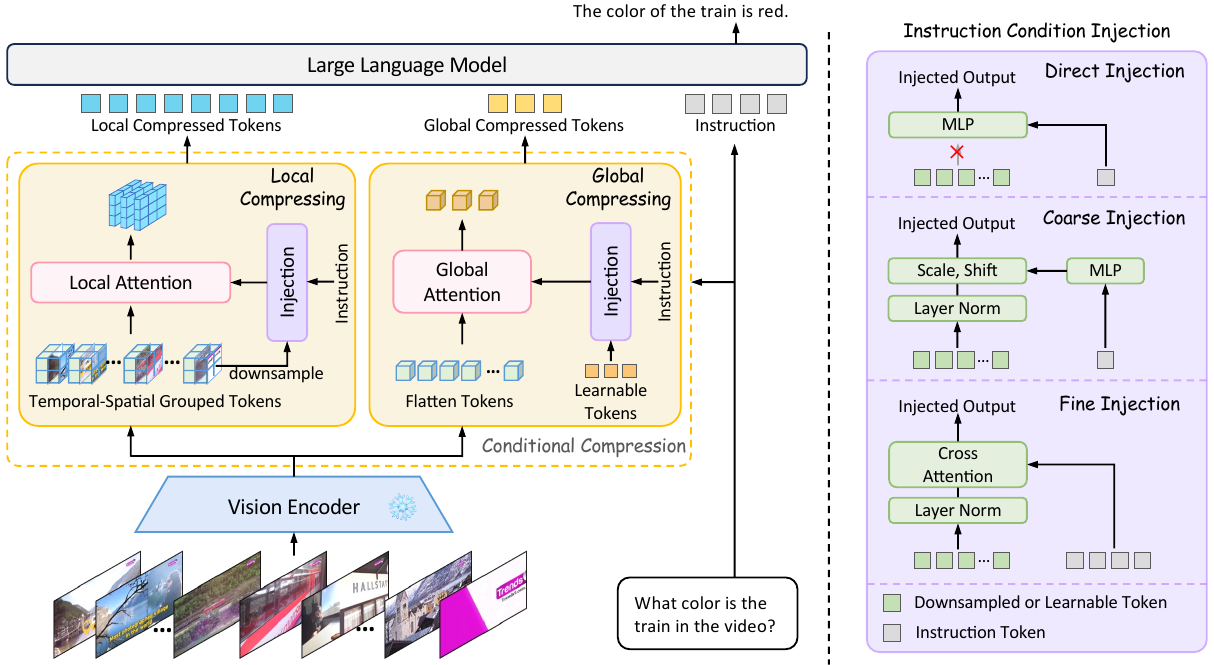}
\caption{The framework of our proposed HICom. We propose the hybrid-level instruction injection to conditionally compress video tokens in MLLMs. We extract instruction-relevant information within each grouped sub-region at the local level, and extract it to a fixed number of tokens at the global level. The instruction condition is injected into the attention process to guide the compression.}
\label{fig:framework}
% \vspace{-8pt}
\end{figure*}

\noindent\textbf{Multi-modal large language models.}
Based on the powerful capabilities of LLMs among various linguistic tasks, many works try to extend the understanding abilities to the computer vision area. Flamingo~\cite{flamingo} and BLIP series~\cite{blip2, instructblip} successfully explore the usage of Resampler and Q-Former to bridge the visual tokens to language models. LLaVA series~\cite{llava, llava1_5, llavanext} use a simple MLP as the visual projector to translate visual tokens to the LLM's embedding space, achieving huge success. Recently, more researchers have focused on extending image tasks to video tasks. The biggest challenge for video tasks is to design an effective method to achieve the trade-off between computational costs and video understanding ability. Early methods~\cite{videollava, videochat, btadapter} sparsely sample frames, lossing too much temporal information. Mainstream methods~\cite{llavanextvideo, llavaov, pllava} use a simple spatial pooling strategy to reduce the number of visual tokens and get a huge success. \cite{videollama, otter} employ Q-Former to compress the spatial-temporal tokens into a fixed number. Chat-Univi~\cite{chatunivi} uses DPC-KNN algorithm to cluster dynamic visual tokens. VideoLLaMA2~\cite{videollama2} utilizes convolution both in temporal and spatial dimensions to downsample tokens. Flash-VStream~\cite{flashvstream} introduces a learnable memory mechanism to compress online video streaming. All of them try their best to observe more visual information. However, their unconditional compression without explicit guidance inherently leads to the loss of instruction-relevant information. Different from them, we conduct conditional compression by injecting instruction at hybrid levels, only emphasizing the visual parts related to the instruction and allowing the irrelevant parts to be discarded, thus getting a better representation for each question.

\noindent\textbf{Text-based visual representation for video LLMs.} There are some methods using instruction information during the visual encoding. \cite{vaquita, yu2024self} sample the frames based on the CLIP~\cite{clip} similarity or a localization model using the instruction, \cite{liang2024end, wang2024weakly} make the frame selection strategy trainable. However, these methods only focus on how to select the correct frames in a rough manner, failing to guide the token compression with the key instruction information.
LLaMA-VID~\cite{llamavid} follows a similar instruction-guided idea, as it introduces the instruction to interact with visual features. However, it roughly simplifies the interacted but uncompressed result into a single token only as auxiliary information for each frame,  destroying the spatial structure and failing to discuss the guiding role of the instruction for conditional compression. Therefore, it is only designed for long videos and the performance is limited.
To fully explore conditional compression, we propose to inject the instruction at hybrid levels, exploring a better way for conditional compression to alleviate the information loss while maintaining the performance.

\section{Method}

\subsection{Instruction-Guided Conditional Compression}
\label{sec:method_comp}
\cref{fig:framework} shows the framework of our proposed HICom, which focuses on the design between the vision encoder and the LLM to conduct the conditional compression. We inject the instruction condition into the compression process at both local and global levels for better information extraction with spatial-temporal structure maintained.

\noindent\textbf{Local-Level Compression.}
Previous works have demonstrated the importance of the spatial inductive bias when processing visual representations for image MLLMs~\cite{cambrian, tokenpacker}. Inspired by them, we divide the tokens into several groups and perform conditional compression within each group to preserve the temporal-spatial structure of video features. 
Specifically, given the encoded video frame features $\boldsymbol{V} \in \mathbb{R}^{T\times H\times W\times D}$, and the downsampling ratio $(\alpha_T, \alpha_H, \alpha_W)$, we divide $\boldsymbol{V}$ into $N_T\cdot$$N_H\cdot$$N_W$ groups, where $N_i=\lceil i / \alpha_i \rceil$, $i \in \{T,H,W\}$. Therefore, there are $\alpha_T\cdot$$\alpha_H\cdot$$\alpha_W$ tokens in each group, which can be formulated as $\boldsymbol{V}^{t,h,w} \in \mathbb{R}^{\alpha_T\times \alpha_H\times \alpha_W\times D}$, where $0 \leq \{t,h,w\} \leq \{N_T,N_H,N_W\}$. We first pool $\boldsymbol{V}^{t,h,w}$ into one token $\boldsymbol{V}_p^{t,h,w} \in \mathbb{R}^{1 \times D}$ and inject the instruction condition $\boldsymbol{C}$ (\ie, the text feature of instructions) into $\boldsymbol{V}_p^{t,h,w}$. Then the local attention within each group can be calculated by:
\begin{equation}
    \boldsymbol{Z}_l^{t,h,w} = \text{Attn}\left[
        \text{Inj}(\boldsymbol{V}_p^{t,h,w}, \boldsymbol{C}), \boldsymbol{V}^{t,h,w}, \boldsymbol{V}^{t,h,w}
    \right],
\label{eq:local_attn}
\end{equation}
where Inj($\cdot$) represents the condition injection, Attn($\cdot$) donates the attention mechanism~\cite{ViT} and the inputs of Attn($\cdot$) are \textit{query}, \textit{key}, and \textit{value}, respectively. Therefore, the visual features within each group are compressed into only one token, preserving the temporal-spatial structure while conditionally highlighting the instruction-relevant parts. Finally, we concatenate $\boldsymbol{Z}_l^{t,h,w}$ as the local compressed results $\boldsymbol{Z}_l \in \mathbb{R}^{N_T \times N_H \times N_W \times D}$, and use an MLP layer to project $\boldsymbol{Z}_l$ into the LLM's embedding space.

\begin{figure}[!t]
\centering
\includegraphics[width=0.42\textwidth]{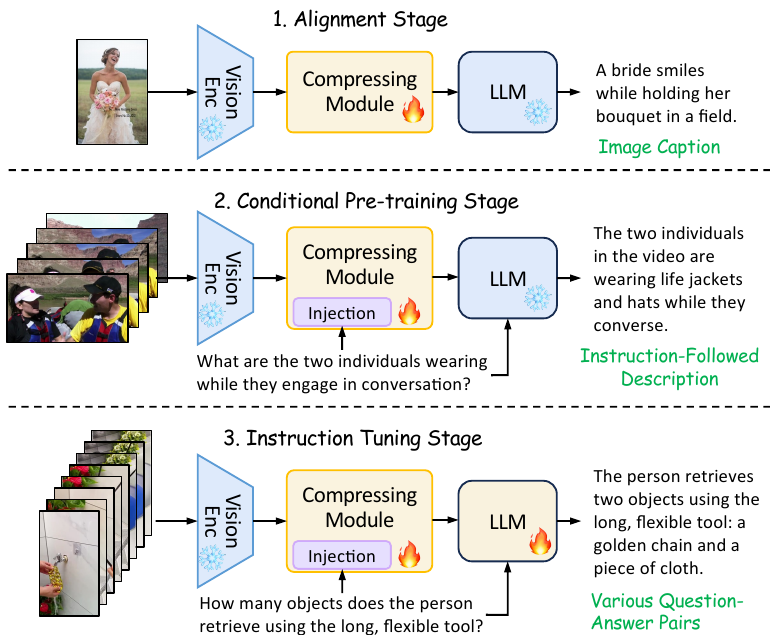}
\caption{We introduce a new guidance pre-training stage and implement three-stage training for conditional compression.}
\label{fig:training_stage}
% \vspace{-8pt}
\end{figure}

\noindent\textbf{Global-Level Compression.}
Though the temporal-spatial structure can be maintained at the local level, the attention is forced to focus on small sub-regions. However, each sub-region may contain information that is not equally valid for the question, and some sub-regions are even quite irrelevant. Therefore, we also implement conditional compression at the global level to highlight the most relevant parts within the whole video.
Specifically, we initialize a small set of learnable tokens $\boldsymbol{L} \in \mathbb{R}^{N_L \times D}$, where $N_L$ donates the number of tokens, and then inject the instruction condition into them. Instead of grouping the video frame features, we apply 3D position embedding Pos($\cdot$) for them and directly flatten them at global-level compression. Then the global compressed tokens $\boldsymbol{Z}_g \in \mathbb{R}^{N_L \times D}$ can be calculated by:
\begin{equation}
    \boldsymbol{Z}_g = \text{Attn}\left[
        \text{Inj}(\boldsymbol{L}, \boldsymbol{C}), \boldsymbol{V} + \text{Pos}(\boldsymbol{V}), \boldsymbol{V}
    \right].
\label{eq:global_attn}
\end{equation}
An MLP layer is also utilized to project $\boldsymbol{Z}_g$ into the LLM's embedding space. And we finally concatenate $\boldsymbol{Z}_l$ and $\boldsymbol{Z}_g$ together for further understanding of LLM.

\noindent\textbf{Instruction Condition Injection.}
In order to fulfill the guiding role of the instruction condition, we explore three different types of injection modules, and define them direct injection, coarse injection, and fine injection, respectively. Note the text encoder can obtain both pooled tokens $\boldsymbol{C}_p \in \mathbb{R}^{1 \times D}$ (\ie, global text embedding) and fine-grained tokens $\boldsymbol{C}_f \in \mathbb{R}^{L \times D}$ (\ie, token-level text embedding). 
Given the pooled token $\boldsymbol{C}_p$, the direct injection uses an MLP to translate the single token into the injected output without any interaction with $\boldsymbol{V}_p^{t,h,w}$ or $\boldsymbol{L}$. We employ coarse injection by the adaptive layer norm~\cite{dit}. Specifically, we use an MLP to regress the scale and shift from $\boldsymbol{C}_p$, then add to the visual input $\boldsymbol{V}_p^{t,h,w}$ or learnable tokens $\boldsymbol{L}$, which we represent them as $\boldsymbol{A}$, after the layer norm. The process can be formulated as:
\begin{equation}
    \text{Inj}(\boldsymbol{A}, \boldsymbol{C}) = \text{LN}(\boldsymbol{A}) \cdot \text{scale}(\boldsymbol{C}_p) + \text{shift}(\boldsymbol{C}_p),
\label{eq:coarse_inj}
\end{equation}
where LN($\cdot$) is the layer norm.
Given the fine-grained tokens $\boldsymbol{C}_f$ as condition $\boldsymbol{C}$, we conduct the fine injection via the cross attention, which can be formulated as:
\begin{equation}
    \text{Inj}(\boldsymbol{A}, \boldsymbol{C}) = \text{Attn}\left(
        \text{LN}(\boldsymbol{A}), \boldsymbol{C}_f,  \boldsymbol{C}_f
    \right).
\label{eq:fine_inj}
\end{equation}
As the direct injection directly translates the condition to the \textit{query} of the attention in \cref{eq:local_attn} and \cref{eq:global_attn}, the token length limits that it is only suitable for local compression as the local branch compresses each group into only one token. On the contrary, the coarse and fine injections are more flexible to fit various situations as the condition is added into $\boldsymbol{A}$.
We conduct the experiments on the three types of injection and finally choose direct injection for local compression and coarse injection for global compression.

\noindent\textbf{Conditional Pre-training Stage.}
The existing mainstream methods follow the pipeline of alignment first and then instruction tuning~\cite{llava, llava1_5, videollama2, flashvstream}, and previous text-based methods~\cite{vaquita, llamavid} also follow this pipeline. However, the image-caption pairs cannot provide valid instruction information in the alignment stage, direct instruction tuning with modules that are not adequately aligned will confuse the guiding process, thus harming the performance.
Therefore, we propose a new conditional pre-training stage between the alignment and the instruction tuning to make the training process easier, as shown in \cref{fig:training_stage}. In this stage, we use our constructed instruction-followed descriptions to pre-train the compression module with condition injection, fulfilling the guiding role of the instruction.

\begin{table}[!t]
\centering
\caption{The statistical information of our constructed HICom-248K dataset.}
\label{tab:dataset_info}
\resizebox{0.4\textwidth}{!}{%
\begin{tabular}{cccc}
\toprule
\textbf{\# Videos}              & \textbf{Sources}                         & \textbf{Avg Length}              & \textbf{\# QA Pairs}            \\ \midrule
\multirow{2}{*}{248K} & Panda70M: 238K                  & \multirow{2}{*}{25.67s} & \multirow{2}{*}{739K} \\
                      & \multicolumn{1}{l}{Ego4D: 10K} &                         &                       \\ \bottomrule
\end{tabular}%
}
\end{table}

\begin{figure}[!t]
\centering
\includegraphics[width=0.4\textwidth]{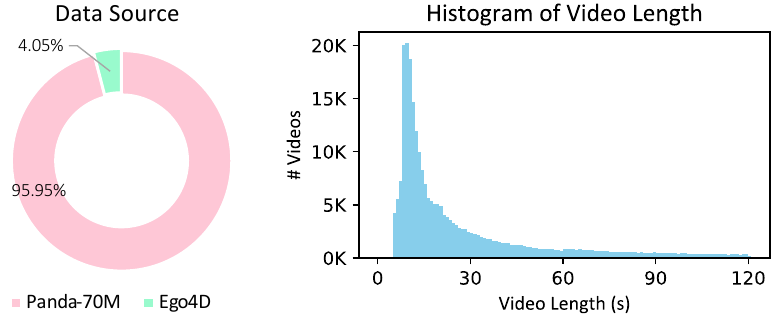}
\caption{The visualization of data source (left) and video length (right) of our constructed HICom-248K dataset.}
\label{fig:dataset_info}
% \vspace{-8pt}
\end{figure}

\begin{table*}[!t]
\setlength\tabcolsep{4pt}
\centering
\caption{Performance comparison between our HICom and other SOTA methods on multiple-choice QA video benchmarks. $\dag$ means we use the result reproduced by VideoLLaMA2~\cite{videollama2}, $*$ indicates we reproduce the results ourselves using the official checkpoint and inference code provided by authors. $\S$ donates we inference with a new length of frames trained by sampling 32 frames.}
\label{tab:mcqa}
\resizebox{0.88\textwidth}{!}{%
\begin{tabular}{@{}lccc|cccccccc|c|c@{}}
\toprule
\multirow{2}{*}{\textbf{Methods}} & \multirow{2}{*}{\textbf{LLM Size}} & \multirow{2}{*}{\textbf{\# Frames}} & \multirow{2}{*}{\textbf{\# Tokens}} & \multicolumn{4}{c}{\textbf{VideoMME w/o sub.}} & \multicolumn{4}{c|}{\textbf{VideoMME w/ sub.}} & \multirow{2}{*}{\textbf{\begin{tabular}[c]{@{}c@{}}MV-\\ Bench\end{tabular}}} & \multirow{2}{*}{\textbf{\begin{tabular}[c]{@{}c@{}}Ego-\\ Schema\end{tabular}}} \\ \cline{5-12}
 &  &  &  & \textbf{short} & \textbf{mid} & \textbf{long} & \textbf{all} & \textbf{short} & \textbf{mid} & \textbf{long} & \textbf{all} &  &  \\ \midrule

Video-LLaVA~\cite{videollava} & 7B & 8 & 2048 & 45.3 & 38.0 & 36.2 & 39.9 & 46.1 & 40.7 & 38.1 & 41.6 & 43.0 & - \\
VideoChat2-Mistral~\cite{mvbench} & 7B & 16 & 1536 & 48.3 & 36.3 & 35.0 & 39.5 & 52.8 & 39.4 & 39.2 & 43.8 & 60.4 & - \\
LLaMA-VID~\cite{llamavid} & 7B & 1fps & 2tps & - & - & - & 25.9$^\dag$ & - & - & - & - & 41.9$^\dag$ & 38.5$^\dag$ \\
Chat-Univi-1.5~\cite{chatunivi} & 7B & 64 & 448 & 45.7 & 40.3 & 35.8 & 40.6 & 51.2 & 44.6 & 41.8 & 45.9 & 45.9 & - \\
PLLaVA~\cite{pllava} & 7B & 16 & 2304 & - & - & - & - & - & - & - & - & 46.6 & - \\
LLaVA-Next-Video~\cite{llavanextvideo} & 7B & 32 & 4608 & 49.9$^*$ & 41.4$^*$ & 37.0$^*$ & 42.8$^*$ & - & - & - & - & 46.5$^\dag$ & 43.9$^\dag$ \\
LongVA~\cite{longva} & 7B & 128 & 18432 & 61.1 & 50.4 & 46.2 & 52.6 & 61.6 & 53.6 & 47.6 & 54.3 & - & - \\
VideoLLaMA2~\cite{videollama2} & 7B & 16 & 1152 & 56.0 & 45.4 & 42.1 & 47.9 & - & - & - & - & 54.6 & 51.7 \\
Tarsier~\cite{tarsier} & 7B & 16 & 2304 & - & - & - & - & - & - & - & - & 62.6 & 56.0 \\
VITA~\cite{vita} & 8$\times$7B & 16 & 4096 & 65.9 & 52.9 & 48.6 & 55.8 & 70.4 & 56.2 & 50.9 & 59.2 & - & - \\
LLaVA-OneVision~\cite{llavaov} & 7B & 32 & 6272 & \underline{70.1}$^*$ & 56.4$^*$ & 48.9$^*$ & 58.5$^*$ & \underline{75.8}$^*$ & 58.4$^*$ & 51.6$^*$ & 61.9$^*$ & 56.7 & 60.1 \\
LLaVA-Video~\cite{llavavideo} & 7B & 32 & 6272 & \textbf{73.9}$^*$ & 57.3$^*$ & 50.4$^*$ & \textbf{60.6}$^*$ & \textbf{76.6}$^*$ & 60.3$^*$ & 51.7$^*$ & 62.9$^*$ & 58.6 & 57.3 \\ \midrule
HICom (Ours) & 1.5B & 32 & 680 & 61.7 & 51.3 & 43.9 & 52.3 & 65.1 & 54.7 & 47.3 & 55.7 & 56.4 & 50.1 \\
HICom (Ours) & 7B & 32 & 680 & 65.7 & 57.1 & \underline{51.4} & 58.1 & 68.4 & 58.9 & 53.1 & 60.1 & \underline{64.1} & 60.5 \\
HICom (Ours)$^{\S}$ & 7B & 64 & 1328 & 67.6 & \underline{58.0} & 51.2 & 58.9 & 71.8 & \underline{63.7} & \underline{54.5} & \underline{63.3} & \textbf{64.7} & \underline{62.2} \\
HICom (Ours)$^{\S}$ & 7B & 128 & 2624 & 69.0 & \textbf{60.2} & \textbf{51.5} & \underline{60.3} & 72.7 & \textbf{64.6} & \textbf{57.3} & \textbf{64.8} & 64.0 & \textbf{62.3} \\ \bottomrule
\end{tabular}%
}
\end{table*}

\subsection{Dataset Construction}
To pre-train the condition injection in the compression module, we construct a new instruction-followed description dataset HICom-248K. Different from common instruction tuning datasets, HICom-248K focuses on providing only one data type, \ie, the description of the visual content related to the instruction, and does not include other rich types of instruction data (\eg,  reasoning, counting, \textit{etc}).

\noindent\textbf{Data Collection.}
To ensure the diversity and quality of our videos, we collect videos from public datasets Panda-70M~\cite{panda70m} and Ego4D~\cite{ego4d}. Panda-70M contains various open-domain videos from YouTube, and Ego4D collects many high-resolution ego-centric human activity videos. 
Notably, we use the original untrimmed videos to cut clips ourselves since the provided split clips of both datasets are too short, greatly decreasing the content complexity.
To balance the number of different types of videos, we pre-define 29 categories~\cite{videomme, llavavideo} using natural language (\eg \textit{A video about cooking activity.}) and use InternVideo2~\cite{internvideo2} to extract both the video features and the pre-defined sentence features. Based on the similarities between them, we select 1,500 videos for each category and then randomly select additional 10,000 videos from the others to ensure diversity.

\noindent\textbf{Video Processing.}
The untrimmed videos are too long for our pre-training, and also severely affect the quality of captions and annotations by existing SOTA models. Therefore, we use PySceneDetect to split each video into shorter ones with a much looser threshold than Panda-70M. As the split clips can sometimes be redundant with repeat semantics, we further extract keyframes for each video and only select the split clips containing the keyframes as our final results. Specifically, inspired by ShareGPT4Video~\cite{sharegpt4video}, we calculate the CLIP score for densely sampled frames within a video, and select the less similar frames as keyframes. Finally, we further filter out the video clips shorter than 5 seconds and longer than 120 seconds. % and get 248K video clips.

\noindent\textbf{Instruction-Followed Description Generation.}
For the processed video clips, we use Qwen2-VL-72B-Instruct~\cite{qwen2vl} to systematically generate the instruction-followed description with the CoT technique~\cite{wei2022chain}. Specifically, we first generate the detailed caption of each input video, then take both the video content and caption as input to generate the three instruction-answer pairs for each video. Different from the normal instruction datasets, we require Qwen2-VL-72B-Instruct to meet the following two rules: 1) The instructions should refer to the specific visual information, such as the people or the objects in the video; 2) The instructions should lead to a descriptive response and the answers should describe the object mentioned in the instruction in detail. Finally, we filter the generated results by discarding answers containing invalid information, such as the phrase "not provide", or "not describe".

\noindent\textbf{Dataset Statistics.}
\cref{tab:dataset_info} and \cref{fig:dataset_info} shows some statistics about our constructed dataset. We collect 248K video clips in total from Panda-70M and Ego4D, the average of the video lengths is 25.67 seconds and the distribution is shown in \cref{fig:dataset_info}. With the help of Qwen2-VL-72B-Instruct, we generate 739K instruction-followed descriptions, providing sufficient data for our guidance pre-training.

\section{Experiments}

\begin{table}[!t]
\setlength\tabcolsep{1pt}
\centering
\caption{Performance of our HICom and other SOTA methods on open-ended QA video benchmarks. All the compared models are 7B. $\dag$, $*$, and $\S$ keep the same meaning with \cref{tab:mcqa}. \textit{\# F.}, \textit{\# T.} are short for \# Frames, \# Tokens, ANet is short for ActivityNet, LLaVA-NV, LLaVA-OV are short for LLaVA-Next-Video, and LLaVA-OneVision. The format we report ANet is Acc/Score.}
\label{tab:oeqa}
\resizebox{0.47\textwidth}{!}{%
\begin{tabular}{@{}lcc|c|cccccc@{}}
\toprule
\multirow{2}{*}{\textbf{Methods}} & \multirow{2}{*}{\textbf{\# F.}} & \multirow{2}{*}{\textbf{\# T.}} & \multirow{2}{*}{\textbf{ANet}} & \multicolumn{6}{c}{\textbf{VideoChatGPT Bench}} \\ \cline{5-10} 
 &  &  &  & \textbf{CI} & \textbf{DO} & \textbf{CU} & \textbf{TU} & \textbf{CO} & \textbf{Avg} \\ \midrule
Video-LLaVA~\cite{videollava} & 8 & 2048 & 45.3/3.3 & - & - & - & - & - & - \\
VideoChat2~\cite{mvbench} & 16 & 1536 & 49.1/3.3 & 3.02 & 2.88 & 3.51 & 2.66 & 2.81 & 2.98 \\
Chat-Univi~\cite{chatunivi} & 64 & 448 & 45.8/3.2 & 2.89 & 2.91 & 3.46 & \underline{2.89} & 2.81 & 2.99 \\
LLaMA-VID~\cite{llamavid} & 1fps & 2tps & 47.4/3.3 & 2.96 & 3.00 & 3.53 & 2.46 & 2.51 & 2.89 \\
LongVLM~\cite{longvlm} & 100 & 305 & 47.6/3.3 & 2.76 & 2.86 & 3.34 & 2.39 & 3.11 & 2.89 \\
ST-LLM~\cite{st_llm} & 16 & 512 & 50.9/3.3 & 3.23 & 3.05 & 3.74 & \textbf{2.93} & 2.81 & 3.15 \\
PLLaVA~\cite{pllava} & 16 & 2304 & 56.3/3.5 & 3.21 & 2.86 & 3.62 & 2.33 & 2.93 & 2.99 \\
LLaVA-NV~\cite{llavanextvideo} & 32 & 4608 & 53.5/3.2 & \underline{3.39} & \textbf{3.29} & \textbf{3.92} & 2.6 & 3.12 & \textbf{3.26} \\
LongVA~\cite{longva} & 128 & 18432 & -/2.8 & 3.05 & \underline{3.09} & \underline{3.77} & 2.44 & \textbf{3.64} & \underline{3.20}\\
VideoLLaMA2~\cite{videollama2} & 16 & 1152 & 50.2/3.3 & 3.16 & 3.08 & 3.69 & 2.56 & 3.14 & 3.13 \\
Tarsier~\cite{tarsier} & 16 & 2304 & \textbf{59.5}/\underline{3.6} & - & - & - & - & - & - \\
SF-LLaVA~\cite{slowfastllava} & 50 & 3680 & 56.3/3.4 & 3.09 & 2.70 & 3.57 & 2.52 & \underline{3.35} & 3.04 \\
LLaVA-OV~\cite{llavaov} & 32 & 6272 & 56.6/\underline{3.6$^*$} & \textbf{3.45}$^*$ & 3.00$^*$ & 3.71$^*$ & 2.68$^*$ & 3.14$^*$ & \underline{3.20$^*$} \\ \midrule
HICom(Ours)-1.5B & 32 & 680 & 53.0/3.5 & 3.09 & 2.67 & 3.40 & 2.41 & 2.98 & 2.91 \\ 
HICom(Ours)-7B & 32 & 680 & 58.3/\textbf{3.7} & 3.29 & 2.85 & 3.59 & 2.67 & 3.22 & 3.12 \\ 
HICom(Ours)-7B$^{\S}$ & 64 & 1328 & \underline{59.4}/\textbf{3.7} & 3.32 & 2.92 & 3.65 & 2.74 & \underline{3.35} & \underline{3.20} \\ 
HICom(Ours)-7B$^{\S}$ & 128 & 2624 & \textbf{59.5}/\textbf{3.7} & 3.33 & 2.86 & 3.65 & 2.74 & 3.32 & 3.18 \\ \bottomrule
\end{tabular}%
}
\end{table}

\subsection{Experimental Setup}

\noindent\textbf{Implementation Details.}
We use SigLIP~\cite{siglip} (so400m-patch14-384) as our vision encoder and text encoder, choose Qwen2.5 series~\cite{qwen2.5} as our LLMs, and randomly initialize the compressor.
The vision encoder keeps frozen at all stages, the LLM is frozen at the two pre-train stages, and is fine-tuned at the instruction tuning stage. We follow LLaVA-OneVision~\cite{llavaov} to choose our training configurations, the global batch size is set to 512 at the alignment stage, 256 at the conditional pre-train and instruction tuning stage. We use the learning rate of 1e-3 at the alignment stage, 1e-5 at the instruction tuning stage. At our proposed conditional pre-training stage, we use 1e-3 for the condition injection sub-module and 1e-4 for other parameters in the compressing module. We train 1 epoch for all stages. More implementation details can be found in \cref{sec:suppl_implementation}.

\noindent\textbf{Datasets.} We use LLaVA-558K~\cite{llava1_5} image-caption pairs for our alignment stage, the constructed 248K HICom-Pretrain instruction-followed descriptions for conditional pre-train stage. Inspired by~\cite{tarsier, llavavideo}, we collect 2.68M video instruction data for instruction tuning, including 1.6M from LLaVA-Video~\cite{llavavideo}, 292K from VideoChat2-IT~\cite{videochat} subset, 255K from M4-Instruct-Data~\cite{llava_next_interleave}, 11K from Charades~\cite{sigurdsson2016hollywood}, 114K from NTU RGB+D~\cite{shahroudy2016ntu}, 122K from TVQA~\cite{lei2018tvqa}, and 290K from MiT~\cite{monfort2019moments}. We evaluate our model on 5 video benchmarks, including 3 multiple-choice benchmarks VideoMME~\cite{videomme}, MVBench~\cite{mvbench}, EgoSchema~\cite{egoschema}, and 2 open-ended benchmarks ActivityNet~\cite{activitynet}, VideoChatGPT Bench~\cite{videochatgpt}.

\begin{table}[!t]
\setlength\tabcolsep{1pt}
\centering
\caption{The ablation study on our proposed HICom, $\alpha_{(T,H,W)}$ donates the downsampling ratio of each dimension, \textit{\# Tok.} donates the number of tokens input to LLMs. 8f means we sample 8 frames for training. The conditional compression combining both local and global achieves the best.}
\label{tab:component}
\resizebox{0.47\textwidth}{!}{%
\begin{tabular}{@{}lcccccccc@{}}
\toprule
\multirow{2}{*}{\textbf{Methods}} & \multirow{2}{*}{$\boldsymbol{\alpha_{(T,H,W)}}$} & \multicolumn{1}{c|}{\multirow{2}{*}{\textbf{\# Tok.}}} & \multicolumn{4}{c|}{\textbf{VideoMME w/o sub.}} & \multicolumn{1}{c|}{\multirow{2}{*}{\textbf{\begin{tabular}[c]{@{}c@{}}MV-\\ Bench\end{tabular}}}} & \multirow{2}{*}{\textbf{\begin{tabular}[c]{@{}c@{}}Ego-\\ Schema\end{tabular}}} \\ \cline{4-7}
 &  & \multicolumn{1}{c|}{} & \textbf{short} & \textbf{mid} & \textbf{long} & \multicolumn{1}{c|}{\textbf{all}} & \multicolumn{1}{c|}{} &  \\ \midrule
\multicolumn{9}{l}{\textcolor{gray}{\ \ \ \textit{Unconditional Compression}}} \\
avg pool & (1,3,3) & \multicolumn{1}{c|}{2592} & 39.3 & 34.9 & 33.0 & \multicolumn{1}{c|}{35.7} & \multicolumn{1}{c|}{44.8} & 43.2 \\
8f, avg pool & (1,3,3) & \multicolumn{1}{c|}{648} & 36.3 & 34.4 & 32.0 & \multicolumn{1}{c|}{34.3} & \multicolumn{1}{c|}{43.6} & 42.7 \\
avg pool & (4,3,3) & \multicolumn{1}{c|}{648} & 36.3 & 33.3 & 32.2 & \multicolumn{1}{c|}{34.0} & \multicolumn{1}{c|}{43.7} & 41.9 \\ \hline
local & (4,3,3) & \multicolumn{1}{c|}{648} & 36.7 & 34.4 & 32.0 & \multicolumn{1}{c|}{34.4} & \multicolumn{1}{c|}{43.7} & \textbf{42.7} \\
global & None & \multicolumn{1}{c|}{32} & 30.0 & 31.6 & 28.9 & \multicolumn{1}{c|}{30.1} & \multicolumn{1}{c|}{34.6} & 33.6 \\
local+global & (4,3,3) & \multicolumn{1}{c|}{680} & \textbf{37.1} & \textbf{34.8} & \textbf{32.3} & \multicolumn{1}{c|}{\textbf{34.7}} & \multicolumn{1}{c|}{\textbf{44.1}} & 42.4 \\ \midrule
\multicolumn{9}{l}{\textcolor{gray}{\ \ \ \textit{Conditional Compression}}} \\
local & (4,3,3) & \multicolumn{1}{c|}{648} & 38.8 & 36.1 & 33.1 & \multicolumn{1}{c|}{36.0} & \multicolumn{1}{c|}{44.0} & 43.2 \\
global & None & \multicolumn{1}{c|}{32} & 30.8 & 30.3 & 29.8 & \multicolumn{1}{c|}{30.3} & \multicolumn{1}{c|}{35.5} & 34.0 \\
local+global & (4,3,3) & \multicolumn{1}{c|}{680} & \textbf{39.0} & \textbf{36.7} & \textbf{34.2} & \multicolumn{1}{c|}{\textbf{36.6}} & \multicolumn{1}{c|}{\textbf{45.0}} & \textbf{43.5} \\ \bottomrule
\end{tabular}%
}
\end{table}

\begin{figure}[!t]
\centering
\includegraphics[width=0.46\textwidth]{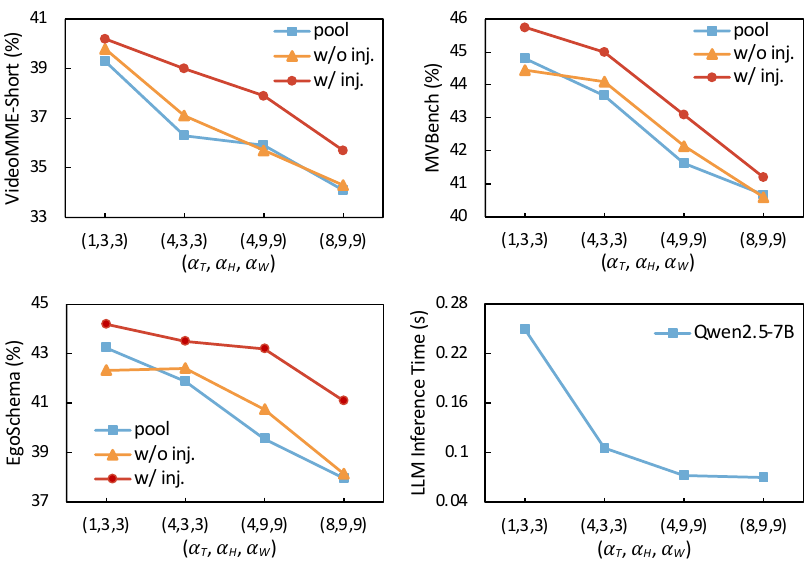}
\caption{The ablation study on different compressing ratios. The figures show the performance on VideoMME-Short (upper left), MVBench (upper right), EgoSchema (lower left), and the inference time of 7B LLM (lower right).}
\label{fig:compress_ratio}
% \vspace{-8pt}
\end{figure}

\subsection{Main Results}
\noindent\textbf{Multiple-choice benchmarks.}
In \cref{tab:mcqa}, we compare the performance of our HICom with different SOTA models on three multiple-choice video benchmarks. We sample 32 frames with 680 tokens to train all our models and sample different frames for inference. Our HICom-7B with 2624 tokens inference achieves the best performance on all three benchmarks. Compared to LLaVA-Video with 6272 tokens, our HICom with only 1328 tokens can obtain better performance, increasing the performance by 2.43\% average and saving 78.8\% tokens, HICom with 2624 tokens gains 3\% averagely when saving 58.2\% tokens, significantly lowering the computational burden. Notably, LLaVA-Video uses numerous additional image data, and it also unfreezes the vision encoder during training, which can both increase performance. It is also noteworthy that though our HICom performs a little worse than LLaVA-Video on VideoMME short videos (less than 2 minutes), we beat LLaVA-Video on both VideoMME medium (4-15 minutes) and long videos (30-60 minutes). We argue that conditional compression is more helpful for long videos. For short videos, the sampled frames are dense enough for unconditional compression, baselines with enough information would naturally perform powerfully. But for long videos, the sampled frames are more sparse with less redundant, our HICom can preserve the maximum amount of instruct-relevant information while it is easily suppressed in LLaVA-Video.

\noindent\textbf{Open-ended benchmarks.}
We also compare the performance of different models on two open-ended video benchmarks, as shown in \cref{tab:oeqa}. We evaluate all our results via GPT-3.5-Turbo-0125. Our HICom also achieves comparable results on both benchmarks, as arriving the SOTA on ActivityNet and getting the second best on the VideoChatGPT Bench. Our HICom with 1328 tokens is on par with Tarsier with much more training data, and performs similarly to LLaVA-OneVision with much more training data and much more visual tokens, demonstrating the effectiveness of our compression.

\noindent\textbf{Generalization Ability on Video Length.} Due to the design of the local-level compression, we can easily extend the lengths of sampled frames at the inference stage for our trained model with 32 frames, rather than re-training it. As we increase the number of sampled frames, the metrics of our HICom also improve across all these benchmarks. For example, the performance on short and medium videos of VideoMME grows fast from 32 frames to 128 frames at our high compressing ratio, as more useful frames are sampled. The performance on long videos of VideoMME without subtitles changes slightly, and we argue this may caused by the extremely long videos (30-60 minutes). Both 32 frames and 128 frames are too sparse for them. Our HICom can pay attention to the most relevant parts based on the sparsely sampled frames, thus changing slightly.

\begin{table}[!t]
\setlength\tabcolsep{4pt}
\centering
\caption{The ablation study on the conditional pre-training stage with the constructed HICom-248K dataset.}
\label{tab:3stage}
\resizebox{0.42\textwidth}{!}{%
\begin{tabular}{@{}lc|cccc|c|c@{}}
\toprule
\multirow{2}{*}{\textbf{Methods}} & \multirow{2}{*}{\textbf{\begin{tabular}[c]{@{}c@{}}Conditional\\ Pre-trained\end{tabular}}} & \multicolumn{4}{c|}{\textbf{VideoMME w/o subs.}} & \multirow{2}{*}{\textbf{\begin{tabular}[c]{@{}c@{}}MV-\\ Bench\end{tabular}}} & \multirow{2}{*}{\textbf{\begin{tabular}[c]{@{}c@{}}Ego-\\ Schema\end{tabular}}} \\ \cline{3-6}
 &  & \textbf{short} & \textbf{mid} & \textbf{long} & \textbf{all} &  &  \\ \midrule
\multirow{2}{*}{avg pool} & \usym{2717} & 36.6 & 32.4 & 33.2 & 34.1 & 42.7 & 40.6 \\
 & \usym{2713} & 36.3 & 33.3 & 32.2 & 34.0 & 43.7 & 41.9 \\ \hline
\multirow{2}{*}{w/o inj.} & \usym{2717} & 39.4 & 33.7 & 33.1 & 35.4 & 42.8 & 41.64 \\
 & \usym{2713} & 37.1 & 34.8 & 32.3 & 34.7 & 44.1 & 42.4 \\ \hline
\multirow{2}{*}{w/ inj.} & \usym{2717} & 38.8 & 35.1 & 34.1 & 36.0 & 44.0 & 41.6 \\
 & \usym{2713} & 39.0 & 36.7 & 34.2 & 36.6 & 45.0 & 43.5 \\ \bottomrule
\end{tabular}%
}
\vspace{-4pt}
\end{table}

\subsection{Ablation Studies}
To save time costs, we choose VideoChat2-IT~\cite{videochat} 896K video data as our instruction tuning data and Qwen2.5-0.5B~\cite{qwen2.5} as our LLM to conduct ablation studies, and extract 32 frames for training unless otherwise specified. This requires less than 28 hours training on 8 A800 GPUs.

\noindent\textbf{Component Analysis.}
The key components of our HICom are local- and global- level compression. \cref{tab:component} shows the results of them in the status of both unconditional (\ie, without condition injection) and conditional compression (\ie, with condition injection). Note we use $\boldsymbol{V}_p^{t,h,w}$ as \textit{query} of \cref{eq:local_attn} in unconditional compression for direct injection. With the same number of input tokens, the unconditional local compression can achieve comparable performance with both spatial pooling (line 2) and temporal-spatial pooling (line 3). Due to the lack of temporal-spatial inductive bias and too few tokens, the global compression significantly harms the performance, and local+global performs only slightly better than local as the additional information of the global branch is limited without explicit guidance. Compared with unconditional compression, the local and global conditional compression increase by 0.8\% and 0.5\%, respectively. The local+global achieves the best, increasing by 1.9\% on VideoMME and 1.3\% average on three benchmarks, also increasing by 0.63\% compared with local conditional compression. As the global branch can provide instruction-relevant information from a global sight, combining both is better. It is noteworthy that our conditional compression with 680 tokens can beat the average pooling with 2592 tokens on all benchmarks, reducing 73.8\% visual tokens and increasing by 0.47\% on the performance.

\begin{table}[!t]
\setlength\tabcolsep{4pt}
\centering
\caption{The ablation study on different types of injection.}
\label{tab:inj}
\resizebox{0.42\textwidth}{!}{%
\begin{tabular}{cc|cccc|c|c}
\toprule
\multirow{2}{*}{\textbf{Local}} & \multirow{2}{*}{\textbf{Global}} & \multicolumn{4}{c|}{\textbf{VideoMME w/o sub.}} & \multicolumn{1}{c|}{\multirow{2}{*}{\begin{tabular}[c]{@{}c@{}}\textbf{MV-}\\ \textbf{Bench}\end{tabular}}} & \multirow{2}{*}{\begin{tabular}[c]{@{}c@{}}\textbf{Ego-}\\ \textbf{Schema}\end{tabular}} \\ \cline{3-6}
 &  & \textbf{short} & \textbf{mid} & \textbf{long} & \textbf{all} &  &  \\ \midrule
direct & \multicolumn{1}{c|}{coarse} & 39.0 & \textbf{36.7} & 34.2 & \textbf{36.6} & \textbf{45.0} & \textbf{43.5} \\
direct & \multicolumn{1}{c|}{fine} & 39.3 & 35.3 & 34.3 & 36.3 & 44.1 & 43.5 \\
coarse & \multicolumn{1}{c|}{coarse} & \textbf{39.8} & 34.8 & \textbf{35.0} & 36.5 & 44.1 & 42.7 \\
coarse & \multicolumn{1}{c|}{fine} & 38.4 & 36.7 & 34.2 & 36.4 & 44.3 & 43.5 \\
fine & \multicolumn{1}{c|}{coarse} & 39.1 & 34.7 & 34.2 & 36.0 & 44.2 & 42.7 \\
fine & \multicolumn{1}{c|}{fine} & 38.8 & 34.9 & 34.2 & 36.0 & 43.6 & 42.0 \\ \bottomrule
\end{tabular}%
}
\vspace{-4pt}
\end{table}

\begin{figure*}[!t]
\centering
\includegraphics[width=0.93\textwidth]{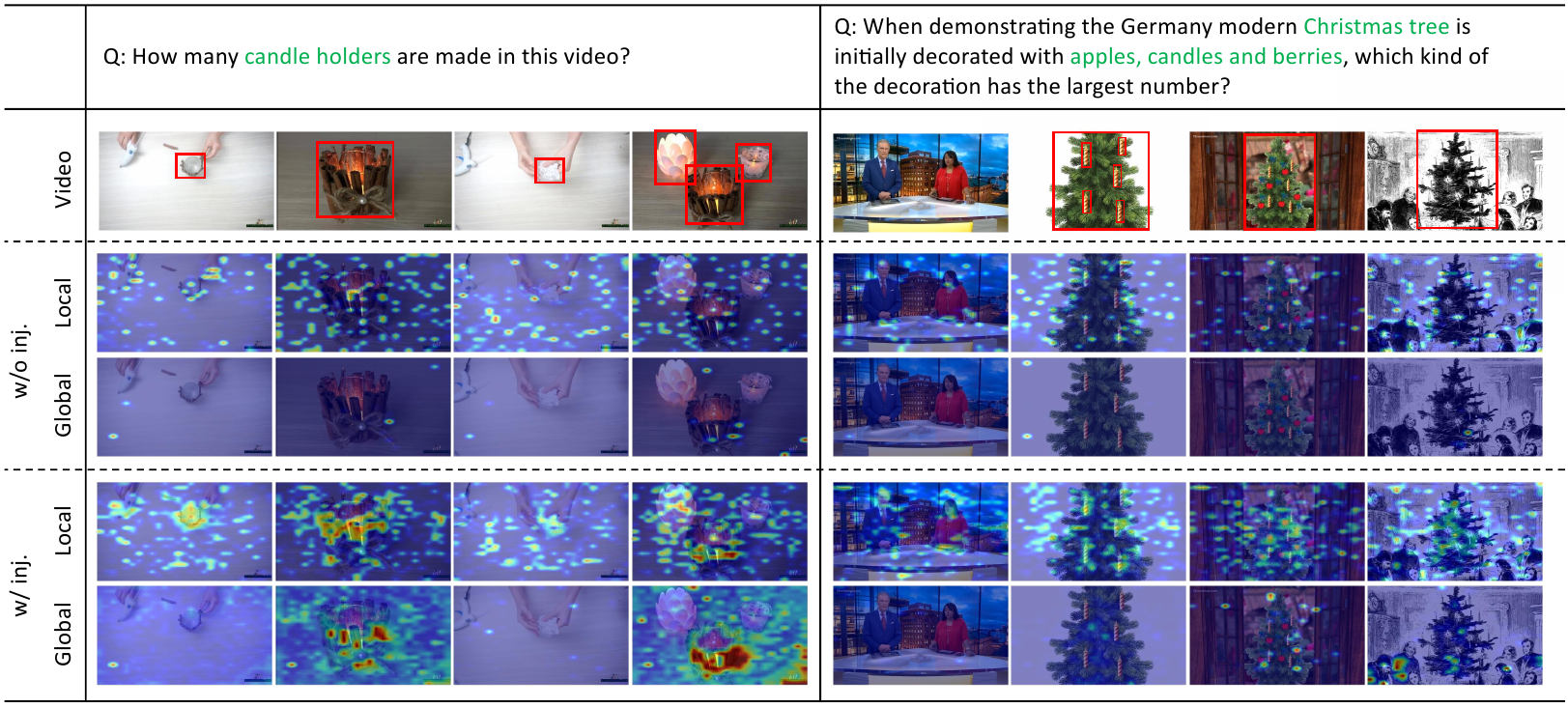}
\caption{The visualization of the attention map at both the local and global level in the situation of unconditional (w/o inj.) and conditional (w/ inj.) compression. We indicate the instruction-relevant parts with the red bounding box for easier reading.}
\label{fig:attention_map}
\vspace{-1pt}
\end{figure*}

\noindent\textbf{Compressing Ratio.}
We explore the effectiveness of the compressing ratio on both unconditional and conditional compression in \cref{fig:compress_ratio}. Our unconditional compression performs similarly with average pooling on three benchmarks at different compressing ratios, and the conditional compression performs much better than them. It is noteworthy that the speed of performance decreasing of the conditional compression is also lower than the unconditional compression as the compressing ratio becomes higher. This shows instruction-relevant information remains with the guidance, demonstrating the superiority. In the lower right sub-figure of \cref{fig:compress_ratio}, the compressing ratio of (4,3,3) saves 57.68\% inference time on Qwen2.5-7B than (1,3,3) with only 2.06\% performance loss on Qwen2.5-0.5B. Higher compressing ratios (4,9,9) and (8,9,9) can only save a little more time with much higher performance loss. Therefore, we choose (4,3,3) as our final compressing ratio.

\noindent\textbf{Conditional Pre-training Stage.}
To evaluate the effectiveness of our proposed conditional pre-training stage, we conduct experiments on whether to use this stage or not on both unconditional and conditional compression in \cref{tab:3stage}. For average pooling and our unconditional compression, the conditional pre-training stage is also effective on MVBench and EgoSchema since more data is trained, but is hard to be effective on VideoMME. For the conditional compression, the conditional pre-training stage is effective on all benchmarks, increasing by 1.17\% averagely, demonstrating the effectiveness of not only the constructed data, but also the newly proposed stage.

\noindent\textbf{Types of Injection.}
We propose three types of instruction condition injection in \cref{sec:method_comp}, direct injection is designed for local compression only, while coarse and fine injection fit both. We conduct the ablation study on them in \cref{tab:inj}. Coarse injection performs better than fine injection on all three benchmarks, and both of them perform better than unconditional compression. We argue the reason may be that the pooled instruction feature is more in line with the CLIP training, and the attention of fine injection is harder to converge than the MLP of coarse injection.

\subsection{Qualitative Analysis}
To qualitatively figure out how our HICom realizes the conditional compression, we visualize the attention map of the compression module in \cref{fig:attention_map}. It clearly shows where the model pays attention to based on the instructions. Since we conduct the local-level compression within the sub-region of each group, the attention map of the local level is more evenly distributed throughout the video than the global level.
Compared to unconditional compression, the attention map of conditional compression is more related to the instruction-relevant visual parts. As shown in \cref{fig:attention_map}, the attention maps at both levels successfully focus on the candle holders in the first example, and successfully on the Christmas tree, the apples, candles, and berries in the second example, which means the model will give more weights of these parts during compressing. We also notice that the global-level compression tends to forget some detailed visual information (\eg, the candle holders in the first and third frame of the first example, the Christmas tree in the second frame of the second example), while the local-level compression is better to capture these details due to group-limited attention.
This shows that hybrid-level compression provides different sights, thus helping the model better understand the videos.

\section{Conclusion}

In this paper, we address the challenge of achieving efficient video understanding in multi-modal large language models (MLLMs) through our proposed HICom. Our approach effectively balances the trade-off between computational costs and information retention by integrating user instructions into the compression process. 
This method not only preserves the temporal-spatial structure of video data but also emphasizes instruction-relevant content through a dual-level compression strategy.
Furthermore, by introducing a conditional pre-training stage with proposed HICom-248K dataset, targeted pre-training enhances the efficacy of instruction-injected schema.
Our experiments demonstrate the effectiveness, as HICom can maintain distinguished video understanding ability with much fewer tokens.
HICom shows a good generalization ability to extend the number of frames during inference, but the fixed frame sampling strategy during training still limits the ability to understand long videos. In the future, we will try to sample frames based on fps and add more long videos for training to further increase the ability. We will also explore the potential of conditional compression on images, especially high-resolution images, making our HICom more powerful.

\section*{Acknowledgment}
This work is supported by the National Key Research and Development Program of China (2022YFB3104700), the National Nature Science Foundation of China (62425114, 62121002, U23B2028, 62232006). We acknowledge the support of Alibaba Group, the GPU cluster built by MCC Lab of Information Science and Technology Institution, USTC, and USTC super-computing center for providing
computational resources for this project.

{
    \small
    \bibliographystyle{ieeenat_fullname}
    \bibliography{main}

\begin{thebibliography}{74}
\providecommand{\natexlab}[1]{#1}
\providecommand{\url}[1]{\texttt{#1}}
\expandafter\ifx\csname urlstyle\endcsname\relax
  \providecommand{\doi}[1]{doi: #1}\else
  \providecommand{\doi}{doi: \begingroup \urlstyle{rm}\Url}\fi

\bibitem[Alayrac et~al.(2022)Alayrac, Donahue, Luc, Miech, Barr, Hasson, Lenc, Mensch, Millican, Reynolds, et~al.]{flamingo}
Jean-Baptiste Alayrac, Jeff Donahue, Pauline Luc, Antoine Miech, Iain Barr, Yana Hasson, Karel Lenc, Arthur Mensch, Katherine Millican, Malcolm Reynolds, et~al.
\newblock Flamingo: a visual language model for few-shot learning.
\newblock \emph{NeurIPS}, 35:\penalty0 23716--23736, 2022.

\bibitem[Bai et~al.(2023)Bai, Bai, Yang, Wang, Tan, Wang, Lin, Zhou, and Zhou]{qwenvl}
Jinze Bai, Shuai Bai, Shusheng Yang, Shijie Wang, Sinan Tan, Peng Wang, Junyang Lin, Chang Zhou, and Jingren Zhou.
\newblock Qwen-vl: A frontier large vision-language model with versatile abilities.
\newblock \emph{arXiv preprint arXiv:2308.12966}, 2023.

\bibitem[Caba~Heilbron et~al.(2015)Caba~Heilbron, Escorcia, Ghanem, and Carlos~Niebles]{activitynet}
Fabian Caba~Heilbron, Victor Escorcia, Bernard Ghanem, and Juan Carlos~Niebles.
\newblock Activitynet: A large-scale video benchmark for human activity understanding.
\newblock In \emph{CVPR}, pages 961--970, 2015.

\bibitem[Chen et~al.(2024{\natexlab{a}})Chen, Wei, Li, Dong, Zhang, Zang, Chen, Duan, Lin, Tang, et~al.]{sharegpt4video}
Lin Chen, Xilin Wei, Jinsong Li, Xiaoyi Dong, Pan Zhang, Yuhang Zang, Zehui Chen, Haodong Duan, Bin Lin, Zhenyu Tang, et~al.
\newblock Sharegpt4video: Improving video understanding and generation with better captions.
\newblock \emph{arXiv preprint arXiv:2406.04325}, 2024{\natexlab{a}}.

\bibitem[Chen et~al.(2024{\natexlab{b}})Chen, Siarohin, Menapace, Deyneka, Chao, Jeon, Fang, Lee, Ren, Yang, et~al.]{panda70m}
Tsai-Shien Chen, Aliaksandr Siarohin, Willi Menapace, Ekaterina Deyneka, Hsiang-wei Chao, Byung~Eun Jeon, Yuwei Fang, Hsin-Ying Lee, Jian Ren, Ming-Hsuan Yang, et~al.
\newblock Panda-70m: Captioning 70m videos with multiple cross-modality teachers.
\newblock In \emph{CVPR}, pages 13320--13331, 2024{\natexlab{b}}.

\bibitem[Chen et~al.(2024{\natexlab{c}})Chen, Wang, Tian, Ye, Gao, Cui, Tong, Hu, Luo, Ma, et~al.]{internvl1_5}
Zhe Chen, Weiyun Wang, Hao Tian, Shenglong Ye, Zhangwei Gao, Erfei Cui, Wenwen Tong, Kongzhi Hu, Jiapeng Luo, Zheng Ma, et~al.
\newblock How far are we to gpt-4v? closing the gap to commercial multimodal models with open-source suites.
\newblock \emph{arXiv preprint arXiv:2404.16821}, 2024{\natexlab{c}}.

\bibitem[Cheng et~al.(2024)Cheng, Leng, Zhang, Xin, Li, Chen, Zhu, Zhang, Luo, Zhao, et~al.]{videollama2}
Zesen Cheng, Sicong Leng, Hang Zhang, Yifei Xin, Xin Li, Guanzheng Chen, Yongxin Zhu, Wenqi Zhang, Ziyang Luo, Deli Zhao, et~al.
\newblock Videollama 2: Advancing spatial-temporal modeling and audio understanding in video-llms.
\newblock \emph{arXiv preprint arXiv:2406.07476}, 2024.

\bibitem[Dai et~al.(2023)Dai, Li, LI, Tiong, Zhao, Wang, Li, Fung, and Hoi]{instructblip}
Wenliang Dai, Junnan Li, DONGXU LI, Anthony Tiong, Junqi Zhao, Weisheng Wang, Boyang Li, Pascale~N Fung, and Steven Hoi.
\newblock Instructblip: Towards general-purpose vision-language models with instruction tuning.
\newblock In \emph{NeurIPS}, pages 49250--49267, 2023.

\bibitem[Dosovitskiy(2020)]{ViT}
Alexey Dosovitskiy.
\newblock An image is worth 16x16 words: Transformers for image recognition at scale.
\newblock \emph{arXiv preprint arXiv:2010.11929}, 2020.

\bibitem[Fu et~al.(2024{\natexlab{a}})Fu, Dai, Luo, Li, Ren, Zhang, Wang, Zhou, Shen, Zhang, et~al.]{videomme}
Chaoyou Fu, Yuhan Dai, Yondong Luo, Lei Li, Shuhuai Ren, Renrui Zhang, Zihan Wang, Chenyu Zhou, Yunhang Shen, Mengdan Zhang, et~al.
\newblock Video-mme: The first-ever comprehensive evaluation benchmark of multi-modal llms in video analysis.
\newblock \emph{arXiv preprint arXiv:2405.21075}, 2024{\natexlab{a}}.

\bibitem[Fu et~al.(2024{\natexlab{b}})Fu, Lin, Long, Shen, Zhao, Zhang, Wang, Yin, Ma, Zheng, et~al.]{vita}
Chaoyou Fu, Haojia Lin, Zuwei Long, Yunhang Shen, Meng Zhao, Yifan Zhang, Xiong Wang, Di Yin, Long Ma, Xiawu Zheng, et~al.
\newblock Vita: Towards open-source interactive omni multimodal llm.
\newblock \emph{arXiv preprint arXiv:2408.05211}, 2024{\natexlab{b}}.

\bibitem[Gao et~al.(2024)Gao, Wang, Qu, Zhang, Wang, Xu, and Xie]{gao2024self}
Zuan Gao, Yuxin Wang, Yadong Qu, Boqiang Zhang, Zixiao Wang, Jianjun Xu, and Hongtao Xie.
\newblock Self-supervised pre-training with symmetric superimposition modeling for scene text recognition.
\newblock \emph{arXiv preprint arXiv:2405.05841}, 2024.

\bibitem[Ge et~al.(2024{\natexlab{a}})Ge, Xie, Li, Xie, Min, and Zhang]{ge2024towards}
Jiannan Ge, Hongtao Xie, Pandeng Li, Lingxi Xie, Shaobo Min, and Yongdong Zhang.
\newblock Towards discriminative feature generation for generalized zero-shot learning.
\newblock \emph{IEEE Transactions on Multimedia}, 2024{\natexlab{a}}.

\bibitem[Ge et~al.(2024{\natexlab{b}})Ge, Xie, Xie, Li, Zhang, Zhang, and Tian]{ge2024alignzeg}
Jiannan Ge, Lingxi Xie, Hongtao Xie, Pandeng Li, Xiaopeng Zhang, Yongdong Zhang, and Qi Tian.
\newblock Alignzeg: Mitigating objective misalignment for zero-shot semantic segmentation.
\newblock In \emph{ECCV}, pages 142--161. Springer, 2024{\natexlab{b}}.

\bibitem[Grauman et~al.(2022)Grauman, Westbury, Byrne, Chavis, Furnari, Girdhar, Hamburger, Jiang, Liu, Liu, et~al.]{ego4d}
Kristen Grauman, Andrew Westbury, Eugene Byrne, Zachary Chavis, Antonino Furnari, Rohit Girdhar, Jackson Hamburger, Hao Jiang, Miao Liu, Xingyu Liu, et~al.
\newblock Ego4d: Around the world in 3,000 hours of egocentric video.
\newblock In \emph{CVPR}, pages 18995--19012, 2022.

\bibitem[He et~al.(2024)He, Li, Jang, Jia, Cao, Shah, Shrivastava, and Lim]{malmm}
Bo He, Hengduo Li, Young~Kyun Jang, Menglin Jia, Xuefei Cao, Ashish Shah, Abhinav Shrivastava, and Ser-Nam Lim.
\newblock Ma-lmm: Memory-augmented large multimodal model for long-term video understanding.
\newblock In \emph{CVPR}, pages 13504--13514, 2024.

\bibitem[Jin et~al.(2024)Jin, Takanobu, Zhang, Cao, and Yuan]{chatunivi}
Peng Jin, Ryuichi Takanobu, Wancai Zhang, Xiaochun Cao, and Li Yuan.
\newblock Chat-univi: Unified visual representation empowers large language models with image and video understanding.
\newblock In \emph{CVPR}, pages 13700--13710, 2024.

\bibitem[Lei et~al.(2018)Lei, Yu, Bansal, and Berg]{lei2018tvqa}
Jie Lei, Licheng Yu, Mohit Bansal, and Tamara~L Berg.
\newblock Tvqa: Localized, compositional video question answering.
\newblock \emph{arXiv preprint arXiv:1809.01696}, 2018.

\bibitem[Li et~al.(2023{\natexlab{a}})Li, Zhang, Chen, Wang, Yang, and Liu]{otter}
Bo Li, Yuanhan Zhang, Liangyu Chen, Jinghao Wang, Jingkang Yang, and Ziwei Liu.
\newblock Otter: A multi-modal model with in-context instruction tuning.
\newblock \emph{arXiv preprint arXiv:2305.03726}, 2023{\natexlab{a}}.

\bibitem[Li et~al.(2024{\natexlab{a}})Li, Zhang, Guo, Zhang, Li, Zhang, Zhang, Li, Liu, and Li]{llavaov}
Bo Li, Yuanhan Zhang, Dong Guo, Renrui Zhang, Feng Li, Hao Zhang, Kaichen Zhang, Yanwei Li, Ziwei Liu, and Chunyuan Li.
\newblock Llava-onevision: Easy visual task transfer.
\newblock \emph{arXiv preprint arXiv:2408.03326}, 2024{\natexlab{a}}.

\bibitem[Li et~al.(2024{\natexlab{b}})Li, Zhang, Zhang, Zhang, Li, Li, Ma, and Li]{llava_next_interleave}
Feng Li, Renrui Zhang, Hao Zhang, Yuanhan Zhang, Bo Li, Wei Li, Zejun Ma, and Chunyuan Li.
\newblock Llava-next-interleave: Tackling multi-image, video, and 3d in large multimodal models.
\newblock \emph{arXiv preprint arXiv:2407.07895}, 2024{\natexlab{b}}.

\bibitem[Li et~al.(2023{\natexlab{b}})Li, Li, Savarese, and Hoi]{blip2}
Junnan Li, Dongxu Li, Silvio Savarese, and Steven Hoi.
\newblock Blip-2: Bootstrapping language-image pre-training with frozen image encoders and large language models.
\newblock In \emph{ICML}, pages 19730--19742. PMLR, 2023{\natexlab{b}}.

\bibitem[Li et~al.(2023{\natexlab{c}})Li, He, Wang, Li, Wang, Luo, Wang, Wang, and Qiao]{videochat}
KunChang Li, Yinan He, Yi Wang, Yizhuo Li, Wenhai Wang, Ping Luo, Yali Wang, Limin Wang, and Yu Qiao.
\newblock Videochat: Chat-centric video understanding.
\newblock \emph{arXiv preprint arXiv:2305.06355}, 2023{\natexlab{c}}.

\bibitem[Li et~al.(2024{\natexlab{c}})Li, Wang, He, Li, Wang, Liu, Wang, Xu, Chen, Luo, et~al.]{mvbench}
Kunchang Li, Yali Wang, Yinan He, Yizhuo Li, Yi Wang, Yi Liu, Zun Wang, Jilan Xu, Guo Chen, Ping Luo, et~al.
\newblock Mvbench: A comprehensive multi-modal video understanding benchmark.
\newblock In \emph{CVPR}, pages 22195--22206, 2024{\natexlab{c}}.

\bibitem[Li et~al.(2023{\natexlab{d}})Li, Xie, Xie, Zhao, Zhang, Zheng, Zhao, and Zhang]{li2023momentdiff}
Pandeng Li, Chen-Wei Xie, Hongtao Xie, Liming Zhao, Lei Zhang, Yun Zheng, Deli Zhao, and Yongdong Zhang.
\newblock Momentdiff: Generative video moment retrieval from random to real.
\newblock \emph{NeurIPS}, 36:\penalty0 65948--65966, 2023{\natexlab{d}}.

\bibitem[Li et~al.(2023{\natexlab{e}})Li, Xie, Zhao, Xie, Ge, Zheng, Zhao, and Zhang]{li2023progressive}
Pandeng Li, Chen-Wei Xie, Liming Zhao, Hongtao Xie, Jiannan Ge, Yun Zheng, Deli Zhao, and Yongdong Zhang.
\newblock Progressive spatio-temporal prototype matching for text-video retrieval.
\newblock In \emph{ICCV}, pages 4100--4110, 2023{\natexlab{e}}.

\bibitem[Li et~al.(2024{\natexlab{d}})Li, Yuan, Liu, Tang, Wang, Zhu, and Zhang]{tokenpacker}
Wentong Li, Yuqian Yuan, Jian Liu, Dongqi Tang, Song Wang, Jianke Zhu, and Lei Zhang.
\newblock Tokenpacker: Efficient visual projector for multimodal llm.
\newblock \emph{arXiv preprint arXiv:2407.02392}, 2024{\natexlab{d}}.

\bibitem[Li et~al.(2024{\natexlab{e}})Li, Wang, and Jia]{llamavid}
Yanwei Li, Chengyao Wang, and Jiaya Jia.
\newblock Llama-vid: An image is worth 2 tokens in large language models.
\newblock In \emph{ECCV}, pages 323--340. Springer, 2024{\natexlab{e}}.

\bibitem[Liang et~al.(2024)Liang, Meng, Wang, Liu, Liu, and Zhao]{liang2024end}
Jianxin Liang, Xiaojun Meng, Yueqian Wang, Chang Liu, Qun Liu, and Dongyan Zhao.
\newblock End-to-end video question answering with frame scoring mechanisms and adaptive sampling.
\newblock \emph{arXiv preprint arXiv:2407.15047}, 2024.

\bibitem[Lin et~al.(2023)Lin, Ye, Zhu, Cui, Ning, Jin, and Yuan]{videollava}
Bin Lin, Yang Ye, Bin Zhu, Jiaxi Cui, Munan Ning, Peng Jin, and Li Yuan.
\newblock Video-llava: Learning united visual representation by alignment before projection.
\newblock \emph{arXiv preprint arXiv:2311.10122}, 2023.

\bibitem[Liu et~al.(2023)Liu, Li, Wu, and Lee]{llava}
Haotian Liu, Chunyuan Li, Qingyang Wu, and Yong~Jae Lee.
\newblock Visual instruction tuning.
\newblock \emph{NeurIPS}, 36, 2023.

\bibitem[Liu et~al.(2024{\natexlab{a}})Liu, Li, Li, and Lee]{llava1_5}
Haotian Liu, Chunyuan Li, Yuheng Li, and Yong~Jae Lee.
\newblock Improved baselines with visual instruction tuning.
\newblock In \emph{CVPR}, pages 26296--26306, 2024{\natexlab{a}}.

\bibitem[Liu et~al.(2024{\natexlab{b}})Liu, Li, Li, Li, Zhang, Shen, and Lee]{llavanext}
Haotian Liu, Chunyuan Li, Yuheng Li, Bo Li, Yuanhan Zhang, Sheng Shen, and Yong~Jae Lee.
\newblock Llava-next: Improved reasoning, ocr, and world knowledge, 2024{\natexlab{b}}.

\bibitem[Liu et~al.(2024{\natexlab{c}})Liu, Li, Ge, Li, Shan, and Li]{btadapter}
Ruyang Liu, Chen Li, Yixiao Ge, Thomas~H Li, Ying Shan, and Ge Li.
\newblock Bt-adapter: Video conversation is feasible without video instruction tuning.
\newblock In \emph{CVPR}, pages 13658--13667, 2024{\natexlab{c}}.

\bibitem[Liu et~al.(2024{\natexlab{d}})Liu, Li, Tang, Ge, Shan, and Li]{st_llm}
Ruyang Liu, Chen Li, Haoran Tang, Yixiao Ge, Ying Shan, and Ge Li.
\newblock St-llm: Large language models are effective temporal learners.
\newblock In \emph{European Conference on Computer Vision}, pages 1--18. Springer, 2024{\natexlab{d}}.

\bibitem[Maaz et~al.(2024)Maaz, Rasheed, Khan, and Khan]{videochatgpt}
Muhammad Maaz, Hanoona Rasheed, Salman Khan, and Fahad~Shahbaz Khan.
\newblock Video-chatgpt: Towards detailed video understanding via large vision and language models.
\newblock In \emph{ACL}, 2024.

\bibitem[Mangalam et~al.(2023)Mangalam, Akshulakov, and Malik]{egoschema}
Karttikeya Mangalam, Raiymbek Akshulakov, and Jitendra Malik.
\newblock Egoschema: A diagnostic benchmark for very long-form video language understanding.
\newblock \emph{NeurIPS}, 36:\penalty0 46212--46244, 2023.

\bibitem[Monfort et~al.(2019)Monfort, Andonian, Zhou, Ramakrishnan, Bargal, Yan, Brown, Fan, Gutfreund, Vondrick, et~al.]{monfort2019moments}
Mathew Monfort, Alex Andonian, Bolei Zhou, Kandan Ramakrishnan, Sarah~Adel Bargal, Tom Yan, Lisa Brown, Quanfu Fan, Dan Gutfreund, Carl Vondrick, et~al.
\newblock Moments in time dataset: one million videos for event understanding.
\newblock \emph{IEEE TPAMI}, 42\penalty0 (2):\penalty0 502--508, 2019.

\bibitem[Peebles and Xie(2023)]{dit}
William Peebles and Saining Xie.
\newblock Scalable diffusion models with transformers.
\newblock In \emph{ICCV}, pages 4195--4205, 2023.

\bibitem[Qu et~al.(2025)Qu, Tang, Peng, Yang, Yu, and Jia]{qu2025doesvisionlanguagemodellost}
Tianyuan Qu, Longxiang Tang, Bohao Peng, Senqiao Yang, Bei Yu, and Jiaya Jia.
\newblock Does your vision-language model get lost in the long video sampling dilemma?, 2025.

\bibitem[Radford et~al.(2021)Radford, Kim, Hallacy, Ramesh, Goh, Agarwal, Sastry, Askell, Mishkin, Clark, et~al.]{clip}
Alec Radford, Jong~Wook Kim, Chris Hallacy, Aditya Ramesh, Gabriel Goh, Sandhini Agarwal, Girish Sastry, Amanda Askell, Pamela Mishkin, Jack Clark, et~al.
\newblock Learning transferable visual models from natural language supervision.
\newblock In \emph{ICML}, pages 8748--8763. PMLR, 2021.

\bibitem[Reid et~al.(2024)Reid, Savinov, Teplyashin, Lepikhin, Lillicrap, Alayrac, Soricut, Lazaridou, Firat, Schrittwieser, et~al.]{gemini1_5}
Machel Reid, Nikolay Savinov, Denis Teplyashin, Dmitry Lepikhin, Timothy Lillicrap, Jean-baptiste Alayrac, Radu Soricut, Angeliki Lazaridou, Orhan Firat, Julian Schrittwieser, et~al.
\newblock Gemini 1.5: Unlocking multimodal understanding across millions of tokens of context.
\newblock \emph{arXiv preprint arXiv:2403.05530}, 2024.

\bibitem[Ren et~al.(2024)Ren, Yao, Li, Sun, and Hou]{timechat}
Shuhuai Ren, Linli Yao, Shicheng Li, Xu Sun, and Lu Hou.
\newblock Timechat: A time-sensitive multimodal large language model for long video understanding.
\newblock In \emph{CVPR}, pages 14313--14323, 2024.

\bibitem[Shahroudy et~al.(2016)Shahroudy, Liu, Ng, and Wang]{shahroudy2016ntu}
Amir Shahroudy, Jun Liu, Tian-Tsong Ng, and Gang Wang.
\newblock Ntu rgb+d: A large scale dataset for 3d human activity analysis.
\newblock In \emph{CVPR}, pages 1010--1019, 2016.

\bibitem[Sigurdsson et~al.(2016)Sigurdsson, Varol, Wang, Laptev, Farhadi, and Gupta]{sigurdsson2016hollywood}
Gunnar~A. Sigurdsson, G{\"u}l Varol, Xiaolong Wang, Ivan Laptev, Ali Farhadi, and Abhinav Gupta.
\newblock Hollywood in homes: Crowdsourcing data collection for activity understanding.
\newblock \emph{ArXiv e-prints}, 2016.

\bibitem[Song et~al.(2024)Song, Chai, Wang, Zhang, Zhou, Wu, Chi, Guo, Ye, Zhang, et~al.]{moviechat}
Enxin Song, Wenhao Chai, Guanhong Wang, Yucheng Zhang, Haoyang Zhou, Feiyang Wu, Haozhe Chi, Xun Guo, Tian Ye, Yanting Zhang, et~al.
\newblock Moviechat: From dense token to sparse memory for long video understanding.
\newblock In \emph{CVPR}, pages 18221--18232, 2024.

\bibitem[Tang et~al.(2024)Tang, Tian, Li, He, Zhou, Zhao, Li, and Jia]{tang2024mind}
Longxiang Tang, Zhuotao Tian, Kai Li, Chunming He, Hantao Zhou, Hengshuang Zhao, Xiu Li, and Jiaya Jia.
\newblock Mind the interference: Retaining pre-trained knowledge in parameter efficient continual learning of vision-language models.
\newblock In \emph{ECCV}, pages 346--365. Springer, 2024.

\bibitem[Team(2024)]{qwen2.5}
Qwen Team.
\newblock Qwen2.5: A party of foundation models, 2024.

\bibitem[Tong et~al.(2024)Tong, Brown, Wu, Woo, Middepogu, Akula, Yang, Yang, Iyer, Pan, et~al.]{cambrian}
Shengbang Tong, Ellis Brown, Penghao Wu, Sanghyun Woo, Manoj Middepogu, Sai~Charitha Akula, Jihan Yang, Shusheng Yang, Adithya Iyer, Xichen Pan, et~al.
\newblock Cambrian-1: A fully open, vision-centric exploration of multimodal llms.
\newblock \emph{arXiv preprint arXiv:2406.16860}, 2024.

\bibitem[Wang et~al.(2024{\natexlab{a}})Wang, Lai, Sun, and Ge]{wang2024weakly}
Haibo Wang, Chenghang Lai, Yixuan Sun, and Weifeng Ge.
\newblock Weakly supervised gaussian contrastive grounding with large multimodal models for video question answering.
\newblock \emph{arXiv preprint arXiv:2401.10711}, 2024{\natexlab{a}}.

\bibitem[Wang et~al.(2024{\natexlab{b}})Wang, Yuan, Zhang, and Sun]{tarsier}
Jiawei Wang, Liping Yuan, Yuchen Zhang, and Haomiao Sun.
\newblock Tarsier: Recipes for training and evaluating large video description models.
\newblock \emph{arXiv preprint arXiv:2407.00634}, 2024{\natexlab{b}}.

\bibitem[Wang et~al.(2024{\natexlab{c}})Wang, Bai, Tan, Wang, Fan, Bai, Chen, Liu, Wang, Ge, et~al.]{qwen2vl}
Peng Wang, Shuai Bai, Sinan Tan, Shijie Wang, Zhihao Fan, Jinze Bai, Keqin Chen, Xuejing Liu, Jialin Wang, Wenbin Ge, et~al.
\newblock Qwen2-vl: Enhancing vision-language model's perception of the world at any resolution.
\newblock \emph{arXiv preprint arXiv:2409.12191}, 2024{\natexlab{c}}.

\bibitem[Wang et~al.(2023)Wang, Zhang, Wang, Bhattacharya, Fu, and Wu]{vaquita}
Yizhou Wang, Ruiyi Zhang, Haoliang Wang, Uttaran Bhattacharya, Yun Fu, and Gang Wu.
\newblock Vaquita: Enhancing alignment in llm-assisted video understanding.
\newblock \emph{arXiv preprint arXiv:2312.02310}, 2023.

\bibitem[Wang et~al.(2024{\natexlab{d}})Wang, Li, Li, Yu, He, Chen, Pei, Zheng, Xu, Wang, et~al.]{internvideo2}
Yi Wang, Kunchang Li, Xinhao Li, Jiashuo Yu, Yinan He, Guo Chen, Baoqi Pei, Rongkun Zheng, Jilan Xu, Zun Wang, et~al.
\newblock Internvideo2: Scaling video foundation models for multimodal video understanding.
\newblock \emph{arXiv preprint arXiv:2403.15377}, 2024{\natexlab{d}}.

\bibitem[Wei et~al.(2022)Wei, Wang, Schuurmans, Bosma, Xia, Chi, Le, Zhou, et~al.]{wei2022chain}
Jason Wei, Xuezhi Wang, Dale Schuurmans, Maarten Bosma, Fei Xia, Ed Chi, Quoc~V Le, Denny Zhou, et~al.
\newblock Chain-of-thought prompting elicits reasoning in large language models.
\newblock \emph{NeurIPS}, 35:\penalty0 24824--24837, 2022.

\bibitem[Wei et~al.(2024{\natexlab{a}})Wei, Zhang, Qing, Yuan, Liu, Liu, Zhang, Zhou, and Shan]{wei2024dreamvideo}
Yujie Wei, Shiwei Zhang, Zhiwu Qing, Hangjie Yuan, Zhiheng Liu, Yu Liu, Yingya Zhang, Jingren Zhou, and Hongming Shan.
\newblock Dreamvideo: Composing your dream videos with customized subject and motion.
\newblock In \emph{CVPR}, pages 6537--6549, 2024{\natexlab{a}}.

\bibitem[Wei et~al.(2024{\natexlab{b}})Wei, Zhang, Yuan, Wang, Qiu, Zhao, Feng, Liu, Huang, Ye, et~al.]{wei2024dreamvideo2}
Yujie Wei, Shiwei Zhang, Hangjie Yuan, Xiang Wang, Haonan Qiu, Rui Zhao, Yutong Feng, Feng Liu, Zhizhong Huang, Jiaxin Ye, et~al.
\newblock Dreamvideo-2: Zero-shot subject-driven video customization with precise motion control.
\newblock \emph{arXiv preprint arXiv:2410.13830}, 2024{\natexlab{b}}.

\bibitem[Weng et~al.(2024)Weng, Han, He, Chang, and Zhuang]{longvlm}
Yuetian Weng, Mingfei Han, Haoyu He, Xiaojun Chang, and Bohan Zhuang.
\newblock Longvlm: Efficient long video understanding via large language models.
\newblock In \emph{ECCV}, pages 453--470. Springer, 2024.

\bibitem[Xu et~al.(2024{\natexlab{a}})Xu, Zhao, Zhou, Lin, Ng, and Feng]{pllava}
Lin Xu, Yilin Zhao, Daquan Zhou, Zhijie Lin, See~Kiong Ng, and Jiashi Feng.
\newblock Pllava: Parameter-free llava extension from images to videos for video dense captioning.
\newblock \emph{arXiv preprint arXiv:2404.16994}, 2024{\natexlab{a}}.

\bibitem[Xu et~al.(2024{\natexlab{b}})Xu, Gao, Gan, Chen, Lai, Gang, Kang, and Dehghan]{slowfastllava}
Mingze Xu, Mingfei Gao, Zhe Gan, Hong-You Chen, Zhengfeng Lai, Haiming Gang, Kai Kang, and Afshin Dehghan.
\newblock Slowfast-llava: A strong training-free baseline for video large language models.
\newblock \emph{arXiv preprint arXiv:2407.15841}, 2024{\natexlab{b}}.

\bibitem[Xu et~al.(2024{\natexlab{c}})Xu, Zhang, Li, Tang, Huang, and Zhang]{xu2024fakeshield}
Zhipei Xu, Xuanyu Zhang, Runyi Li, Zecheng Tang, Qing Huang, and Jian Zhang.
\newblock Fakeshield: Explainable image forgery detection and localization via multi-modal large language models.
\newblock \emph{arXiv preprint arXiv:2410.02761}, 2024{\natexlab{c}}.

\bibitem[Yang et~al.(2024)Yang, Teng, Zheng, Ding, Huang, Xu, Yang, Hong, Zhang, Feng, et~al.]{cogvideox}
Zhuoyi Yang, Jiayan Teng, Wendi Zheng, Ming Ding, Shiyu Huang, Jiazheng Xu, Yuanming Yang, Wenyi Hong, Xiaohan Zhang, Guanyu Feng, et~al.
\newblock Cogvideox: Text-to-video diffusion models with an expert transformer.
\newblock \emph{arXiv preprint arXiv:2408.06072}, 2024.

\bibitem[Yu et~al.(2023)Yu, Cho, Yadav, and Bansal]{yu2024self}
Shoubin Yu, Jaemin Cho, Prateek Yadav, and Mohit Bansal.
\newblock Self-chained image-language model for video localization and question answering.
\newblock \emph{NeurIPS}, 36, 2023.

\bibitem[Zhai et~al.(2023)Zhai, Mustafa, Kolesnikov, and Beyer]{siglip}
Xiaohua Zhai, Basil Mustafa, Alexander Kolesnikov, and Lucas Beyer.
\newblock Sigmoid loss for language image pre-training.
\newblock In \emph{ICCV}, pages 11975--11986, 2023.

\bibitem[Zhang et~al.(2024{\natexlab{a}})Zhang, Xie, Gao, and Wang]{zhang2024choose}
Boqiang Zhang, Hongtao Xie, Zuan Gao, and Yuxin Wang.
\newblock Choose what you need: Disentangled representation learning for scene text recognition removal and editing.
\newblock In \emph{CVPR}, pages 28358--28368, 2024{\natexlab{a}}.

\bibitem[Zhang et~al.(2025)Zhang, Li, Cheng, Hu, Yuan, Chen, Leng, Jiang, Zhang, Li, et~al.]{videollama3}
Boqiang Zhang, Kehan Li, Zesen Cheng, Zhiqiang Hu, Yuqian Yuan, Guanzheng Chen, Sicong Leng, Yuming Jiang, Hang Zhang, Xin Li, et~al.
\newblock Videollama 3: Frontier multimodal foundation models for image and video understanding.
\newblock \emph{arXiv preprint arXiv:2501.13106}, 2025.

\bibitem[Zhang et~al.(2023)Zhang, Li, and Bing]{videollama}
Hang Zhang, Xin Li, and Lidong Bing.
\newblock Video-llama: An instruction-tuned audio-visual language model for video understanding.
\newblock \emph{arXiv preprint arXiv:2306.02858}, 2023.

\bibitem[Zhang et~al.(2024{\natexlab{b}})Zhang, Wang, Tang, Liu, Feng, Dai, and Jin]{flashvstream}
Haoji Zhang, Yiqin Wang, Yansong Tang, Yong Liu, Jiashi Feng, Jifeng Dai, and Xiaojie Jin.
\newblock Flash-vstream: Memory-based real-time understanding for long video streams.
\newblock \emph{arXiv preprint arXiv:2406.08085}, 2024{\natexlab{b}}.

\bibitem[Zhang et~al.(2024{\natexlab{c}})Zhang, Zhang, Li, Zeng, Yang, Zhang, Wang, Tan, Li, and Liu]{longva}
Peiyuan Zhang, Kaichen Zhang, Bo Li, Guangtao Zeng, Jingkang Yang, Yuanhan Zhang, Ziyue Wang, Haoran Tan, Chunyuan Li, and Ziwei Liu.
\newblock Long context transfer from language to vision.
\newblock \emph{arXiv preprint arXiv:2406.16852}, 2024{\natexlab{c}}.

\bibitem[Zhang et~al.(2024{\natexlab{d}})Zhang, Li, Yu, Xu, Li, and Zhang]{zhang2024editguard}
Xuanyu Zhang, Runyi Li, Jiwen Yu, Youmin Xu, Weiqi Li, and Jian Zhang.
\newblock Editguard: Versatile image watermarking for tamper localization and copyright protection.
\newblock In \emph{CVPR}, pages 11964--11974, 2024{\natexlab{d}}.

\bibitem[Zhang et~al.(2024{\natexlab{e}})Zhang, Li, Liu, Lee, Gui, Fu, Feng, Liu, and Li]{llavanextvideo}
Yuanhan Zhang, Bo Li, haotian Liu, Yong~jae Lee, Liangke Gui, Di Fu, Jiashi Feng, Ziwei Liu, and Chunyuan Li.
\newblock Llava-next: A strong zero-shot video understanding model, 2024{\natexlab{e}}.

\bibitem[Zhang et~al.(2024{\natexlab{f}})Zhang, Wu, Li, Li, Ma, Liu, and Li]{llavavideo}
Yuanhan Zhang, Jinming Wu, Wei Li, Bo Li, Zejun Ma, Ziwei Liu, and Chunyuan Li.
\newblock Video instruction tuning with synthetic data.
\newblock \emph{arXiv preprint arXiv:2410.02713}, 2024{\natexlab{f}}.

\bibitem[Zhou et~al.(2024)Zhou, Shu, Zhao, Wu, Xiao, Yang, Xiong, Zhang, Huang, and Liu]{mlvu}
Junjie Zhou, Yan Shu, Bo Zhao, Boya Wu, Shitao Xiao, Xi Yang, Yongping Xiong, Bo Zhang, Tiejun Huang, and Zheng Liu.
\newblock Mlvu: A comprehensive benchmark for multi-task long video understanding.
\newblock \emph{arXiv preprint arXiv:2406.04264}, 2024.

\bibitem[Zhu et~al.(2023)Zhu, Chen, Shen, Li, and Elhoseiny]{minigpt4}
Deyao Zhu, Jun Chen, Xiaoqian Shen, Xiang Li, and Mohamed Elhoseiny.
\newblock Minigpt-4: Enhancing vision-language understanding with advanced large language models.
\newblock \emph{arXiv preprint arXiv:2304.10592}, 2023.

\end{thebibliography}
}
\clearpage
\setcounter{page}{1}
\setcounter{table}{0}
\setcounter{figure}{0}
\renewcommand{\thetable}{A\arabic{table}}
\renewcommand{\thefigure}{A\arabic{figure}}
\appendix
\maketitlesupplementary

\noindent\textbf{Overview.} In the Supplementary Material, we introduce more implementation details in \cref{sec:suppl_implementation}, more details about the conditional pre-training stage and the constructed HICom-248K dataset in \cref{sec:suppl_new_data_stage}. Then we add more experiments in \cref{sec:suppl_more_exp}, including more ablation studies, more benchmarks, and more qualitative analysis.

\section{Implementation Details}
\label{sec:suppl_implementation}

We use SigLIP~\cite{siglip} (so400m-patch14-384) as our vision encoder and text encoder, and choose Qwen2.5 series~\cite{qwen2.5} as our LLMs. At the global-level, the number of the learnable queries is set to 32 referred to Qformer~\cite{blip2}.
We mainly follow LLaVA-OneVision~\cite{llavaov} to configure our training hyper-parameters and settings. The detailed configurations are shown in \cref{tab:train_setting}. Different from LLaVA-OneVision, we keep our vision encoder frozen at all stages. Compared with the \textit{Stage1.5} of LLaVA-OneVision, the training goal of our conditional pre-training stage is quite different. Therefore, we use a few different training strategies. Specifically, we keep the LLM frozen at the conditional pre-training stage as it is designed to align the compressing module based on the instruction condition. We also use a higher learning rate at this stage, 1e-3 for the parameters of instruction injection, and 1e-4 for other parameters in the compressing module.

We use 3D position embedding for our global-level instruction injection, as the input includes three dimensions: time, width, and height. Following CogVideoX~\cite{cogvideox} and Qwen2VL~\cite{qwen2vl}, we extend 2D absolute position embedding to 3D. Each latent in the video tensor can be represented by a 3D
coordinate. We occupy 3/8, 3/8, and 2/8 of the hidden states’ channel of position embedding. The resulting encoding is
then concatenated along the channel dimension to obtain the final 3D positional encoding.

\begin{table}[!t]
\setlength\tabcolsep{4pt}
\centering
\caption{Detailed training configurations for each stage. We follow LLaVA-OneVision~\cite{llavaov} to choose our configurations. At the conditional pre-training stage and instruction tuning stage, we use a global batch size of 512 for the 0.5B model, and 256 for the 7B and 72B models. Comp. denotes our compressing module, which plays the role of compressing the visual tokens and projecting them into the LLM's embedding space.}
\label{tab:train_setting}
\resizebox{0.46\textwidth}{!}{%
\begin{tabular}{@{}l|ccc@{}}
\toprule
 & \textbf{Alignment} & \textbf{\begin{tabular}[c]{@{}c@{}}Conditional\\ Pre-train\end{tabular}} & \textbf{\begin{tabular}[c]{@{}c@{}}Instruction\\ Tuning\end{tabular}} \\ \midrule
\textbf{Data} & Image & Video & Video \\
\# Tokens & 81+32 & 648+32 & 648+32 \\
\# Samples & 558K & 248K & 2.6M \\ \midrule
\textbf{Trainable} & Comp. & Comp. & Comp., LLM \\
7B LLM & 63M & 63M & 7.7B \\ \midrule
\textbf{Batch size} & 512 & 256/512 & 256/512 \\
\textbf{lr}: Vision Enc. & - & - & - \\
\textbf{lr}: inj. In Comp. & - & $1 \times 10^{-3}$ & $1 \times 10^{-5}$ \\
\textbf{lr}: others in Comp. & $1 \times 10^{-3}$ & $1 \times 10^{-4}$ & $1 \times 10^{-5}$ \\
\textbf{lr}: LLM & - & - & $1 \times 10^{-5}$ \\
\textbf{Epoch} & 1 & 1 & 1 \\ \bottomrule
\end{tabular}%
}
\end{table}

\section{New Dataset and Training Stage}
\label{sec:suppl_new_data_stage}

\begin{figure*}[!t]
\centering
\includegraphics[width=0.96\textwidth]{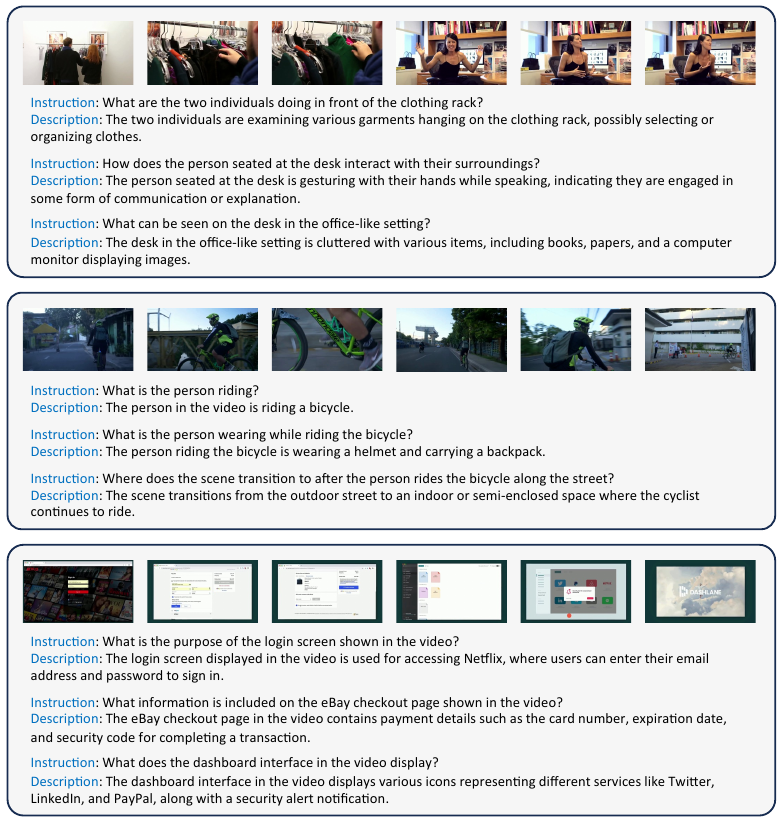}
\caption{Some examples of our constructed HICom-248K instruction-following descriptions.}
\label{fig:dataset_example}
% \vspace{-8pt}
\end{figure*}

\subsection{Conditional Pre-training Stage}
LLaVA~\cite{llava} introduces a two-stage training pipeline for MLLMs, which first pre-trains for feature alignment and then conducts end-to-end instruction tuning. Mainstream methods typically adopt this two-stage training pipeline. During the alignment stage, image-caption pairs are commonly used to pre-train the visual projector, aligning visual features with the LLM's embedding space. At the instruction tuning stage, various types of question-answer pair data are utilized to fine-tune the model, including general QA, multiple-choice QA, OCR, documents/charts/screens, math reasoning, attribute perception, counting, temporal reasoning, information synthesis, \textit{etc}~\cite{llavanext}, equipping the model with instruction-following capabilities.
Recently, LLaVA-OneVision~\cite{llavaov} proposes a three-stage training paradigm. Between \textit{Language-Image Alignment} and \textit{Visual Instruction Tuning}, it introduces a new \textit{stage1.5}, named \textit{High-Quality Knowledge Learning}, using re-captioned detailed description data, OCR data, and Chinese and language data, maintaining a training strategy similar to the instruction tuning stage to inject new knowledge into MLLMs. 

In our work, we propose a new training stage between alignment and instruction tuning. Unlike LLaVA-OneVision, our conditional pre-training stage is designed to pre-train the parameters for instruction injection. In the existing alignment stage, instructions typically prompt the model to describe visual content globally (e.g., "Please summarize the visual content of the image."), which is insufficient for selectively retaining information during token compression. Pre-training the instruction injection at this stage does not effectively teach the model to retain user-important information. By using conditional captions with appropriate instructions, we can pre-train the compression module to perform conditional alignment first, simplifying the instruction tuning for conditional compression. Thus, we introduce a new conditional pre-training stage utilizing our HICom-248K dataset, which implements conditional pre-training for conditional compression.

\begin{table}[!t]
\centering
\caption{The pre-defined 29 categories during the collection of videos in HICom-248K.}
\label{tab:dataset_category}
\resizebox{0.47\textwidth}{!}{%
\begin{tabular}{l}
\toprule
\textbf{Categories defined with natural language} \\ \midrule
A video about cooking activity. \\
A video about writing activity. \\
A video about travel. \\
A video about sight-seeing activity. \\
A life record video about exercise. \\
A life record video about daily life. \\
A life record video about handcraft. \\
A life record video about food. \\
A TV news report video. \\
A video about computer games. \\
A video about sports. \\
A video about football. \\
A video about basketball. \\
A video about pets and animals. \\
A video about action movie scene. \\ 
A video about comedy movie scene. \\
A video about sci-fi movie scene. \\
A video about crime movie scene. \\
A video about horror movie scene. \\
A video about magic show. \\
A video about acrobatics. \\
A documentary or TV show about humanity and history. \\
A documentary or TV show about biography. \\
A documentary or TV show about geography. \\
A documentary or TV show about finance and commerce. \\
A documentary or TV show about literature and art. \\
A documentary or TV show about biology and medicine knowledge. \\
A documentary or TV show about finance and commerce knowledge. \\
A documentary or TV show about technology knowledge. \\ \bottomrule
\end{tabular}%
}
\end{table}

\subsection{HICom-248K}
HICom-248K dataset is designed for the conditional pre-training, which consists of video question-answer pairs. Since the goal of the conditional pre-training stage is to achieve conditional alignment based on the instruction, HICom-248K focuses on providing one type of data, \ie, the instruction-followed descriptions, which meets the following requests:
\begin{itemize}
    \item The instruction should refer to the specific information in the video, providing the guidance role of conditional compression.
    \item The answer should be the caption of the specific visual content of the video mentioned in the instruction.
\end{itemize}
We collect the videos from Panda-70M~\cite{panda70m} and Ego4D~\cite{ego4d}. To ensure the diversity of the video sources, we pre-define 29 categories~\cite{llavavideo, videomme} using natural language, select 1,500 videos for each category based on the similarity score calculated by InternVideo2~\cite{internvideo2}, and randomly select additional 10,000 videos from the whole Panda-70M and Ego4D datasets. The 29 categories are shown in \cref{tab:dataset_category}.
\cref{fig:dataset_example} shows some examples of our constructed HICom-248K. We use the open-soured Qwen2-VL-72B-Instruct~\cite{qwen2vl} to generate around three instruction-description pairs for each video. The generated descriptions follow the instructions well and also accurately capture the visual content, which is suitable for conditional pre-training.

\begin{table}[!t]
\setlength\tabcolsep{3pt}
\centering
\caption{The ablation study on the group strategy for the local-level compression.}
\label{tab:group}
\resizebox{0.46\textwidth}{!}{%
\begin{tabular}{@{}lc|cccc|c|c@{}}
\toprule
\multirow{2}{*}{\textbf{Methods}} & \multirow{2}{*}{\textbf{w/ group}} & \multicolumn{4}{c|}{\textbf{VideoMME w/o sub.}} & \multirow{2}{*}{\textbf{\begin{tabular}[c]{@{}c@{}}MV-\\ Bench\end{tabular}}} & \multirow{2}{*}{\textbf{\begin{tabular}[c]{@{}c@{}}Ego-\\ Schema\end{tabular}}} \\ \cline{3-6}
 &  & \textbf{short} & \textbf{mid} & \textbf{long} & \textbf{all} &  &  \\ \midrule
\multirow{2}{*}{Unconditional} & \usym{2713} & 36.7 & 34.4 & 32 & 34.4 & 43.7 & 42.7 \\
 & \usym{2717} & 34.7 & 31.9 & 31 & 32.5 & 42.9 & 39.9 \\ \midrule
\multirow{2}{*}{Conditional}  & \usym{2713} & 38.8 & 36.1 & 33.1 & 36.0 & 44.0 & 43.2 \\
 & \usym{2717} & 36.6 & 33.7 & 31.2 & 33.8 & 43.6 & 41.6 \\ \bottomrule
\end{tabular}%
}
\end{table}

\begin{table}[!t]
\centering
\caption{The ablation study on valid and invalid instruction on VideoMME without subtitles. We manually select 326 samples with invalid instructions and 2374 samples with valid instructions.}
\label{tab:valid_ins}
\resizebox{0.42\textwidth}{!}{%
\begin{tabular}{@{}lc|cccc@{}}
\toprule
\multicolumn{2}{l|}{\textbf{Methods}} & \textbf{Short} & \textbf{Medium} & \textbf{Long} & \textbf{All} \\ \midrule
\multicolumn{1}{l|}{\multirow{3}{*}{Valid}} & \# Samples & 808 & 816 & 750 & 2374 \\
\multicolumn{1}{l|}{} & w/o inj. & 34.1 & 33.5 & 31.1 & 32.9 \\
\multicolumn{1}{l|}{} & w/ inj. & 36.3 & 35.5 & 33.2 & 35.0 \\ \midrule
\multicolumn{1}{l|}{\multirow{3}{*}{Invalid}} & \# Samples & 92 & 84 & 150 & 326 \\
\multicolumn{1}{l|}{} & w/o inj. & 63.0 & 47.6 & 38.7 & 47.9 \\
\multicolumn{1}{l|}{} & w/ inj. & 63.0 & 47.6 & 39.3 & 48.1 \\ \bottomrule
\end{tabular}%
}
\end{table}

\begin{table}[!t]
\setlength\tabcolsep{2pt}
\centering
\caption{Ablation study about the Conditional Pre-training stage (CP for short) and HICom-248K data with different training strategies. We keep the projector of LLaVA-OV/LLaVA-Video (\ie, two layers of MLP, 2$\times$2 spatial pooling) to train a baseline with our ablation data. We report the result on Video-MME without subtitles and EgoSchema.}
\label{tab:cp_stage}
\resizebox{0.46\textwidth}{!}{%
\begin{tabular}{@{}lccc@{}}
\toprule
\textbf{Traning Strategy} & \textbf{Methods} & \textbf{VideoMME} & \textbf{EgoSchema} \\ \midrule
\multirow{2}{*}{2 Stage w/o CP} & Baseline & 36.1 & 42.5 \\
 & HICom & 36.0 & 41.6 \\ \midrule
\multirow{2}{*}{2 Stage mix HICom-248K for SFT} & Baseline & 36.4 & 43.3 \\
 & HICom & 36.2 & 42.4 \\ \midrule
\multirow{2}{*}{3 Stage w/ HICom-248K for CP} & Baseline & 36.2 & 43.2 \\
 & HICom & \textbf{36.6} & \textbf{43.5} \\ \midrule
3 Stage w/ random 248K MCQA for CP & HICom & 34.6 & 40.8 \\ \bottomrule
\end{tabular}%
}
\end{table}

\section{More Experiments}
\label{sec:suppl_more_exp}

\subsection{More Ablation Studies}
We implement more ablation studies here to demonstrate the superiority and generalization ability of our HICom.

\noindent\textbf{The group strategy at the local level.}
We introduce the temporal-spatial inductive bias, group the visual tokens, and conduct the local-level conditional compression within each group, preserving the temporal-spatial structure while highlighting the instruction-relevant visual content. We evaluate this grouping strategy for the local-level compression in \cref{tab:group}. Without the grouping strategy, the performance drops significantly, especially on VideoMME and EgoSchema benchmarks, showing the significance of explicitly maintaining the temporal-spatial structure.

\begin{figure*}[!t]
\centering
\includegraphics[width=\textwidth]{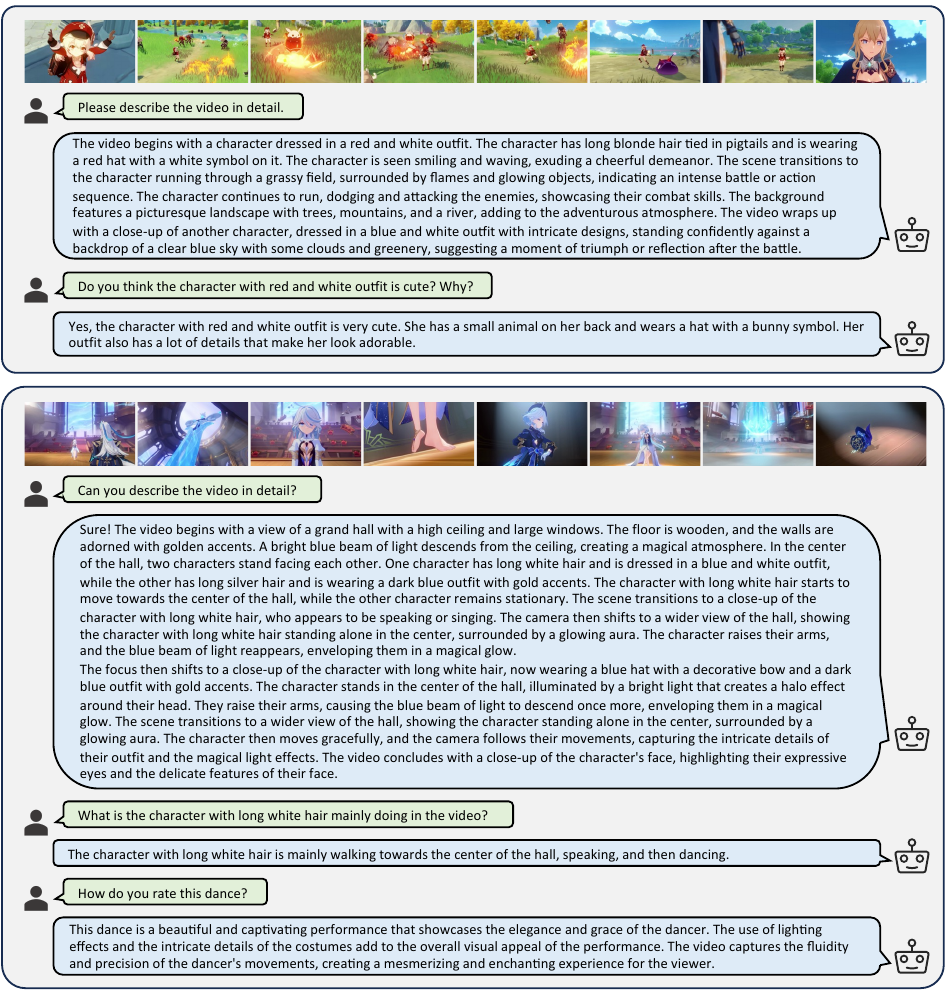}
\caption{Some video dialogue examples of HICom in the scene of the animated style.}
\label{fig:genshin}
% \vspace{-8pt}
\end{figure*}

\noindent\textbf{Valid and invalid instruction.}
We notice that not all instructions can provide effective guidance information for capturing visual information, \eg, the instruction of the caption task. We call them the invalid instruction. To evaluate the performance of our HICom in this situation, we manually select out 326 samples with invalid instructions in the VideoMME benchmark. We list some examples of our selected invalid instruction as follows:
\begin{itemize}
    \item What is this video mainly about?
    \item What can be learned from this video?
    \item Which element doesn't show up in the video?
    \item In what order were the following mentioned in the video?
    \item According to the video, which of the following statements is true?
    \item Which of the following accurately describes the content of the video?
\end{itemize}

\begin{figure*}[!h]
\centering
\includegraphics[width=\textwidth]{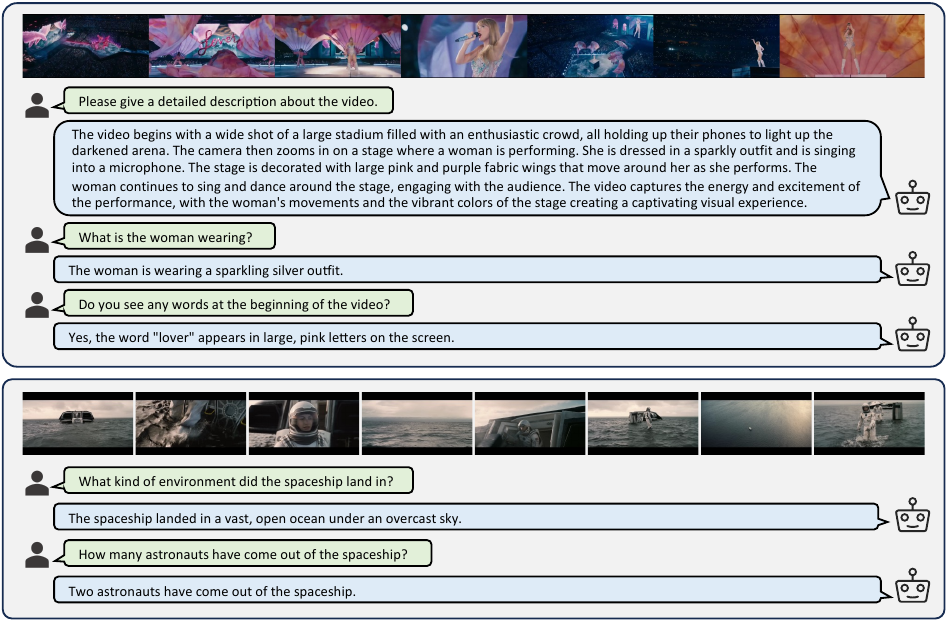}
\caption{Some video dialogue examples of HICom in the scene of the realistic style.}
\label{fig:dialogue2}
% \vspace{-8pt}
\end{figure*}

\begin{table}[!t]
\setlength\tabcolsep{2pt}
\centering
\caption{Inference efficiency comparison between LLaVA-OneVision-7B and our HICom-7B. We report the number of parameters, the inference time of each component, and the final throughput.}
\label{tab:throughput}
\resizebox{0.46\textwidth}{!}{%
\begin{tabular}{@{}clccccc@{}}
\toprule
 & \textbf{Methods} & \textbf{Frames} & \textbf{Vision Encoder} & \textbf{Compressor} & \textbf{LLM} & \textbf{All} \\ \midrule
\multirow{2}{*}{Params} & LLaVA-OV-7B & 32 & 413M & 16M & 7.6B & 8.0B \\
 & HICom-7B & 32 & 428M & 450M+63M & 7.6B & 8.5B \\ \midrule
\multirow{2}{*}{\begin{tabular}[c]{@{}c@{}}Time\\ (ms)\end{tabular}} & LLaVA-OV-7B & 32 & 11.1 & 2.3 & 553.7 & 567.1 \\
 & HICom-7B & 32 & 11.1 & 23.9 & 102.7 & 137.7 \\ \midrule
\multirow{4}{*}{\begin{tabular}[c]{@{}c@{}}Throughput\\ (s/video)\end{tabular}} & LLaVA-OV-7B & 32 & - & - & - & 4.25 \\
 & HICom-7B & 32 & - & - & - & 1.51 \\
 & HICom-7B & 64 & - & - & - & 1.89 \\
 & HICom-7B & 128 & - & - & - & 2.68 \\ \bottomrule
\end{tabular}%
}
\end{table}

\begin{table}[!t]
\centering
\caption{The comparison of SOTA methods and HICom on MLVU$_{dev}$ benchmark. $^*$ indicates we reproduce the results ourselves using the official checkpoint and inference code provided by authors. $\S$ donates we inference with a new length of frames trained by sampling 32 frames.}
\label{tab:mlvu}
\resizebox{0.47\textwidth}{!}{%
\begin{tabular}{@{}lcccc@{}}
\toprule
\textbf{Methods} & \textbf{LLM Size} & \textbf{Frames} & \textbf{Tokens} & \textbf{MLVU$_{dev}$} \\ \midrule
Video-LLaVA~\cite{videollava} & 7B & 8 & 2048 & 47.3 \\
LLaMA-VID~\cite{llamavid} & 7B & 1fps & 2tps & 33.2 \\
LongVA~\cite{longva} & 7B & 128 & 18432 & 56.3 \\
VideoLLaMA2~\cite{videollama2} & 7B & 16 & 1152 & 48.5 \\
LLaVA-OneVision~\cite{llavaov} & 7B & 32 & 6272 & 65.3$^*$ \\
LLaVA-Video~\cite{llavavideo} & 7B & 32 & 6272 & \underline{66.7}$^*$ \\ \midrule
HICom (Ours) & 7B & 32 & 680 & 62.8 \\
HICom (Ours)$^\S$ & 7B & 64 & 1328 & 65.1 \\
HICom (Ours)$^\S$ & 7B & 128 & 2624 & \textbf{67.2} \\ \bottomrule
\end{tabular}%
}
\end{table}

As shown in \cref{tab:valid_ins}, we test both the unconditional and conditional compression on 326 invalid instructions and the other 2374 valid instructions of VideoMME without subtitles separately. For valid instructions, the conditional compression (w/ inj.) gains 2.1\% compared with unconditional compression (w/o inj.). When it comes to invalid instructions, the conditional compression keeps the same results as unconditional compression on short and medium videos, and even gains slightly on long videos. We also find the performance of this situation might be easier than valid instruction, as both models perform much better. Thanks to our design of the local-level compression and the group strategy, we argue that the conditional compression will also focus on the global content of the video within each group, and degrade to the situation of unconditional compression, as there also exists this kind of data during training. The conditional compression will not perform lower than the unconditional compression.

\noindent\textbf{Conditional pre-training stage and HICom-248K data.} We further conduct ablation studies on our proposed conditional pre-training stage and HICom-248K data. We use four different training strategy settings, and report their results of both baseline and our HICom, as shown in\cref{tab:cp_stage}. We keep the projector of LLaVA-OV/LLaVA-Video (\ie, two layers of MLP, 2$\times$2 spatial pooling) to train a baseline with our ablation data. We find the increase from constructed data for baseline (\ie, the comparison between the first strategy and the third strategy) is not as significant as HICom (averagely 0.4\% vs 1.25\%). We also find that for HICom, mixing SFT (the second strategy) gains slightly (0.5\%) on 2-stage training (the first strategy), but our 3-stage training (the third strategy) outperforms 2-stage training (the first strategy) obviously (1.25\%). These two findings demonstrate that our improvement comes more from the additional pre-training strategy, rather than the constructed data themselves. The fourth strategy demonstrates the significance of the data type of the conditional stage, as the performance significantly drops when we change our instruction-followed descriptions to multi-choice QA data, which may confuse the alignment of instruction injection.

\noindent\textbf{Inference efficiency.} Apart from the number of visual tokens that are sent into LLM, we also report the throughput to further demonstrate the inference efficiency. The comparison between LLaVA-OneVision-7B and HICom-7B is shown in \cref{tab:throughput}. We report the time using the same sample, and we only report the LLM time of the first generated token for fair comparison. We report the average time-consuming result of inferring 100 samples as throughput, the time in throughput is larger than our reported time because the throughput counts the time of video loading, the {\tt prepare\_inputs\_labels\_for\_multimodal} process, and the LLM generates more than one token. Compared with LLaVA-OneVision, our vision encoder includes an additional projector and therefore contains additional 16M parameters, and the compressor includes an additional 450M text encoder. This leads to our 23.9ms consumption of the compressor with the text encoding process, 21.6ms more than LLaVA-OneVision. However, our compressor significantly reduces the visual tokens, resulting much shorter time consumption of LLM (102.7ms \textit{vs} 553.7ms), as the LLM inference time usually occupies the main part. Therefore, the number of visual tokens can also effectively and accurately reflect the inference efficiency. Thanks to our compression, our final throughput is much faster than LLaVA-OneVision, as our HICom with 128 frames still infers 1.6x faster than LLaVA-OneVision with only 32 frames.

\subsection{More Benchmarks}
We also report our HICom on MLVU~\cite{mlvu} in \cref{tab:mlvu} as MLVU is a long video benchmark. We report the performance of MLVU's dev split on multi-choice tasks. Our HICom also achieves comparable performance among SOTA 7B models.

\subsection{More Qualitative Analysis}
We provide some examples of video dialogues in this section to intuitively show the video understanding ability. \cref{fig:genshin} shows the situation of animated style videos. Thanks to the design of the local-level compression, our HICom also obtains powerful caption capabilities and gives detailed and accurate captions for videos. Meanwhile, HICom can make judgments based on the understanding of the video. For example, it captures the girl's outfit to draw the conclusion that the girl is cute in the first example video, and think the dancing is beautiful and captivating after watching the second example video. \cref{fig:dialogue2} shows the scene of realistic style videos, which is also easy for HICom to handle. In the first example video, HICom accurately describes the concert scene, and also recognize the word "Lover" on the large screen. In the second movie clip example, HICom can perceive the environment information and accurately count how many people are in the scene.


\end{document}